%% file: emnlp2023.tex
\pdfoutput=1

\documentclass[11pt]{article}

\usepackage[]{EMNLP2023}

\usepackage{times}
\usepackage{latexsym}

\usepackage[T1]{fontenc}

\usepackage[utf8]{inputenc}

\usepackage{microtype}

\usepackage{inconsolata}

\newcommand{\gnof}{\textsc{905}\xspace}
\newcommand{\gacorncourt}{\textsc{acorncourt}\xspace}
\newcommand{\gadvent}{\textsc{advent}\xspace}
\newcommand{\gadventureland}{\textsc{adventureland}\xspace}
\newcommand{\ganchor}{\textsc{anchor}\xspace}
\newcommand{\gawaken}{\textsc{awaken}\xspace}
\newcommand{\gbalances}{\textsc{balances}\xspace}
\newcommand{\gdeephome}{\textsc{deephome}\xspace}
\newcommand{\gdetective}{\textsc{detective}\xspace}
\newcommand{\gdragon}{\textsc{dragon}\xspace}
\newcommand{\genchanter}{\textsc{enchanter}\xspace}
\newcommand{\ginhumane}{\textsc{inhumane}\xspace}
\newcommand{\gjewel}{\textsc{jewel}\xspace}
\newcommand{\gkarn}{\textsc{karn}\xspace}
\newcommand{\glibrary}{\textsc{library}\xspace}
\newcommand{\gludicorp}{\textsc{ludicorp}\xspace}
\newcommand{\gmoonlit}{\textsc{moonlit}\xspace}
\newcommand{\gomniquest}{\textsc{omniquest}\xspace}
\newcommand{\gpentari}{\textsc{pentari}\xspace}
\newcommand{\greverb}{\textsc{reverb}\xspace}
\newcommand{\gsnacktime}{\textsc{snacktime}\xspace}
\newcommand{\gsorcerer}{\textsc{sorcerer}\xspace}
\newcommand{\gspellbrkr}{\textsc{spellbrkr}\xspace}
\newcommand{\gspirit}{\textsc{spirit}\xspace}
\newcommand{\gtemple}{\textsc{temple}\xspace}
\newcommand{\gzenon}{\textsc{zenon}\xspace}
\newcommand{\gzorko}{\textsc{zork1}\xspace}
\newcommand{\gzorkt}{\textsc{zork3}\xspace}
\newcommand{\gztuu}{\textsc{ztuu}\xspace}

\usepackage{amsmath} 
\usepackage{amssymb} 
\usepackage{amsfonts} 
\usepackage{mathrsfs} 
\usepackage{graphicx} 
\usepackage{soul} 
\usepackage{bm}
\usepackage{xspace}
\usepackage{multicol}
\usepackage{multirow}
\usepackage{tabularx}
\usepackage{booktabs}
\usepackage{arydshln}
\usepackage{subcaption}
\usepackage{color} 
\usepackage{xcolor} 
\usepackage{ragged2e,array}
\usepackage{tabularx}
\usepackage{booktabs}
\usepackage{multirow}
\usepackage{multicol}
\usepackage{booktabs}
\usepackage{ragged2e,array,booktabs}
\usepackage{graphicx}
\usepackage{arydshln}
\usepackage{listings} 

\newcommand{\stitle}[1]{\vspace{0.3ex} \noindent{\bf #1}}

\title{A Minimal Approach for Natural Language Action Space\\in Text-based Games}

\author{
  Dongwon Kelvin Ryu$^\spadesuit$
  \ \ \ \ \ 
  Meng Fang$^\heartsuit$
  \ \ \ \ \ 
  Gholamreza Haffari$^\spadesuit$
  \\
  \textbf{Shirui Pan}$^\diamondsuit$
  \ \ \ \ \
  \textbf{Ehsan Shareghi}$^\spadesuit$ $^\clubsuit$
  \\
  $^\spadesuit$~Department of Data Science \& AI, Monash University
  \\
  $^\heartsuit$~University of Liverpool\ \ \ \ 
  $^\diamondsuit$~Griffith University
  \\
  $^\clubsuit$~Language Technology Lab, University of Cambridge
  \\
  \texttt{firstname.lastname@monash.edu}\ \ \ \ 
  \texttt{Meng.Fang@liverpool.ac.uk}
  \\
  \texttt{s.pan@griffith.edu.au}
}

\begin{document}
\maketitle
\input{section/sec-0-abstract.tex}

\input{section/sec-1-introduction.tex}

\input{section/sec-2-related-work.tex}

\input{section/sec-4-text-based-actor-critic.tex}
\input{section/sec-5-experiments}

\input{section/sec-6-discussion}

\input{section/sec-7-conclusion}

\bibliography{emnlp2023}
\bibliographystyle{acl_natbib}

\clearpage

\appendix

\input{figure/tac-decoder}

\section*{Appendices}
\label{sec:appendix}

{
In this section, we provide the details of TAC, training, and full experimental results. We also provide Limitations and Ethical Considerations. 
}


\section{Hyperparameters} \label{appendix:hyper-parameters}

{
Table \ref{tab:hyper-parameters} shows the hyper-parameters used for our experiments.
For \gnof, \gadvent, \ganchor, \gawaken, \gdeephome, \ginhumane and \gmoonlit, gradients exploding has been observed with the hyper-parameters in Table \ref{tab:hyper-parameters}, so we reduced learning rate to $ 10^{-5} $ for these games.
}

\input{table/tab-hyper-parameters}


\section{Parameter Size for \gzorko} \label{appendix:parameter-size}

{
The total parameter size of TAC in \gzorko is 1,783,849 with 49,665 target state critic, which slightly varies by the size of template and object space per game.
This is much lower than KG-A2C (4,812,741), but little higher than DRRN (1,486,081).\footnote{The code for KG-A2C is in \url{https://github.com/rajammanabrolu/KG-A2C}, and DRRN is in \url{https://github.com/microsoft/tdqn}.}
}

\input{table/tab-parameter-size-zork1}


\section{Training Time} \label{appendix:training-time}

{
We used Intel Xeon Gold 6150 2.70 GHz for CPU and Tesla V100-PCIE-16GB for GPU, 8 CPUs with 25GB memory, to train KG-A2C and TAC on \gzorko.
The results are demonstrated in Table \ref{tab:training-time-hyper-parameters}.\footnote{The code for KG-A2C is in \url{https://github.com/rajammanabrolu/KG-A2C}.}
Our TAC has approximately three times lesser parameters than KG-A2C in \gzorko, which would be consistent across different games.
On the other hand, for step per second, TAC is twice faster in GPU and thrice faster in CPU than KG-A2C.
Approximated days for training TAC on CPU and GPU are 1.2 and 0.8 days while KG-A2C is 4.1 and 1.6 days.
TAC still benefits from GPU, but not as much as KG-A2C as its training time is more dependent to the game engine than back-propagation.
}

\input{table/tab-trn-time}


\section{Details of Actor and Critic Components} \label{appendix:actor-critic}
{
Consider an action example (\texttt{take OBJ from OBJ}, \texttt{egg}, \texttt{fridge}) as (template, first object, second object).
Template $ a_{ \mathbb{T} } = $ (\texttt{take OBJ from OBJ}) is sampled from template decoder and encoded to $ h_{ \mathbb{T} } $ with text encoder.
Object decoder takes action representation $ a $ and encoded semi-completed action $ h_{ \mathbb{T} } $ and produces the first object $ a_{ \mathbb{O}1 } = $ (\texttt{egg}).
The template $ a_{ \mathbb{T} } = $ (\texttt{take OBJ from OBJ}) and the first object $ a_{ \mathbb{O}1 } = $ (\texttt{egg}) are combined to $ a_{ \mathbb{T} , \mathbb{O}1 } = $ (\texttt{take egg from OBJ}), $ a_{ \mathbb{T} } \otimes a_{ \mathbb{O}1 } = a_{ \mathbb{T} , \mathbb{O}1 } $.
$ a_{ \mathbb{T} , \mathbb{O}1 } $ is then, encoded to hidden state $ h_{ \mathbb{T}, \mathbb{O}1 } $ with text encoder.
Similarly, the object decoder takes $ a $ and $ h_{ \mathbb{T} , \mathbb{O}1 } $ and produces the second object $ a_{ \mathbb{O}2 } = $ (\texttt{fridge}).
$ a_{ \mathbb{T} , \mathbb{O}1 } $ and $ a_{ \mathbb{O}2 } $ are combined to be natural language action, $ a_{ \mathbb{T} , \mathbb{O}1 } \otimes a_{ \mathbb{O}2 } = a_{ N } $
Finally, $ a_{N} $ is encoded to $ h_{a} $ with text encoder and inputted to state-action critic to predict Q value.
}


\input{table/tab-qual1-case}

\input{table/tab-qual1-anal}

\section{Comparison with Vanilla A2C in \citet{kg-a2c-trl-jericho}} \label{appendix:comparison-kg-a2c}

\paragraph{Architecture.}
{
Vanilla A2C from \citet{kg-a2c-trl-jericho} uses separate gated recurrent units (GRUs) to encode textual observations and previous action, $ ( o_{\text{game}} , o_{\text{look}} , o_{\text{inv}} , a_{t-1} ) $, and transforms the game score, $ n_{\text{score}} $, into binary encoding.
Then, they are concatenated and passed through state network to form state representation.
Their state network is GRU-based to account historical information.
The actor-critic network consists of actor and state value critic, so the state representation is used to estimate state value and produce the policy distribution.
}

{
Our TAC uses a single shared GRU to encode textual observations and previous action with different initial state to signify that the text encoder constructs the general representation of text while the game score is embedded to learnable high dimentional vector.
However, when constructing state representation, we only used $ ( o_{\text{game}} , o_{\text{look}} , o_{\text{inv}} ) $ under our observation that $ o_{\text{game}} $ carries semantic information about $ a_{t-1} $.
Additionally, we also observed that the learned game score representation acts as conditional vector in Appendix \ref{appendix:qualitative-analysis}, so the state representation is constructed as an instance of observation without historical information.
Finally, we included additional modules, state-action value critic \citep{sac-rl-robotics}, target state critic \citep{ddqn-rl-atari} and two state-action critics \citep{two-critics-rl, sac-rl-robotics} for practical purpose.
}

\paragraph{Objective Function.}
{
Three objectives are employed in \citet{kg-a2c-trl-jericho}, reinforcement learning (RL), supervised learning (SL) and entropy regularization.
Both RL and SL are also used in our objectives with minor changes in value function update in RL.
That is, two state-action value critics are updated independently to predict Q value per state-action pair and target state critic is updated as moving average of state critic
Notable difference is that we excluded entropy regularization from \citet{kg-a2c-trl-jericho}.
This is because under our ablation in Section 5.2, we observed that SL acts as regularization.
}

\paragraph{Replay Buffer}
{
Unlike on-policy vanilla A2C \citep{kg-a2c-trl-jericho}, since TAC utilizes $ \epsilon $-admissible exploration, it naturally sits as off-policy algorithm.
We used prioritized experience replay (PER) as our replay buffer \citep{per}.
Standard PER assigns a newly acquired experience with the maximum priority.
This enforces the agent to prioritize not-yet-sampled experiences over others.
As we are using 32 parallel environments and 64 batch size for update, half of the updates will be directed by newly acquired experiences, which not all of them may be useful.
Thus, instead, we assign newly acquired experience with TD errors when they are added to the buffer.
This risks the agent not using some experiences, but it is more efficient since we sample useful batch of experiences.
}


\input{table/tab-qual2-case}

\input{table/tab-qual2-anal}

\section{Qualitative Analysis} \label{appendix:qualitative-analysis}

{
It has been repetitively reported that including game score when constructing state helps in TGs \cite{kg-a2c-trl-jericho, mc-lave-trl-jericho-mcts-rave}.
Here, we provide some insights in what the agent learns from the observations using fully trained TAC.
To illustrate this, we highlight the role of game score on the action preference of the TAC for the same observation in \gzorko.
Observations for different cases can be found in Table \ref{tab:qual1-case} and Table \ref{tab:qual2-case} while the policy and Q value are in Table \ref{tab:qual1-anal} and Table \ref{tab:qual2-anal}.
}

\paragraph{Case 1 in Table \ref{tab:qual1-case} and Table \ref{tab:qual1-anal}}
{
For three different cases, \texttt{Case 1.1}, \texttt{Case 1.2}, and \texttt{Case 1.3}, the agent is at \texttt{Kitchen} location, so many semantic meaning between textual observations are similar, i.e. $ o_{ \text{look} } $ or $ o_{ \text{inv} } $.
For each case, the agent is meant to go \texttt{west} with $ n_{ \text{score} } = 10 $, go \texttt{west} with $ n_{ \text{score} } = 39 $, and go \texttt{east} with $ n_{ \text{score} } = 45 $, respectively.
In \texttt{Case 1.1}, despite the optimal choice of action is \texttt{west}, by replacing the score from $ n_{ \text{score} } = 10 $ to $ n_{ \text{score} } = 45 $, the agent chooses \texttt{east}, which is appropriate for \texttt{Case 1.3}.
Another interesting observation is that replacing game score decreases Q value from 23.7460 to 5.0134 for \texttt{west} and from 18.4385 to 6.0319 for \texttt{east} in \texttt{Case 1.1}.
This seems like the agent thinks it has already acquired reward signals between $ n_{ \text{score} } = 10 $ and $ n_{ \text{score} } = 45 $, resulting in a reduction in Q value.
We speculate that this is because the embedding of $ n_{ \text{score} } $ carries some inductive bias, i.e. temporal, for the agent to infer the stage of the game.
This is consistently manifested in \texttt{Case 1.3}, but in \texttt{Case 1.2}, the agent is robust to the game score because it carries \texttt{painting} that is directly related to reward signals, navigating to pursue that particular reward, which is \texttt{put paining in case} for reward signal of $ +6 $ in \texttt{Living Room} location.
}

\input{figure/tac-main}

\input{table/tab-main-all}

\paragraph{Case 2 in Table \ref{tab:qual2-case} and Table \ref{tab:qual2-anal}}
{
In \texttt{Case 2}, the agent is at \texttt{Behind House} for three other sets of game instances, which has action and score pair as, \texttt{open window} for $ n_{ \text{score} } = 0 $, \texttt{west} for $ n_{ \text{score} } = 0 $, and \texttt{north} for $ n_{ \text{score} } = 45 $.
The phenomenon between \texttt{Case 1.1} and \texttt{Case 1.3} occurs the same for \texttt{Case 2.2} and \texttt{Case 2.3}.
However, unlike \texttt{Case 1}, the score between \texttt{Case 2.1} and \texttt{Case 2.2} is the same.
This means that the agent somehow chooses the optimal action for \texttt{Case 2.2} over \texttt{Case 2.1} in the case where $ n_{ \text{score} } = 0 $ is injected for \texttt{Case 2.3}.
This appears to be that the agent can capture semantic correlation between \texttt{"In one corner of the house there is a small window which is open"} from textual observation in \texttt{Case 2.3} and \texttt{open window} action.
Because \texttt{a small window} is already opened, \texttt{open window} action is no longer required, so the agent tends to produce \texttt{west}, which is appropriate for \texttt{Case 2.2}.
}

{
Thus, from our qualitative analysis, we speculate that the agent captures the semantics of the textual observations and infers the game stage from game score embedding to make optimal decision. 
}


\section{Full Experimental Results} \label{appendix:full-results}

{
The full learning curve of TAC and game score comparison are presented in Figure \ref{fig:tac-main} and Table \ref{tab:main-all}.
}


\section{Stronger Supervised Signals for \gzorko} \label{appendix:stronger-supervised-signals}

{
We also explored how stronger supervised signals can induce better regularization in \gzorko.
Similar to other sets of experiments, we selected variety of $ \lambda_{ \mathbb{T} } $-$ \lambda_{ \mathbb{O} } $ pair.
However, our results show that TAC starts under-fitting in \gzorko when larger $ \lambda_{ \mathbb{T} } $ and $ \lambda_{ \mathbb{O} } $ are applied.
}

\input{figure/tac-regularization-zork1}


\input{figure/tac-dynamic-eps-1.tex}

\input{figure/tac-dynamic-eps-2.tex}

\section{Adaptive Score-based $\epsilon$} \label{appendix:adaptive}
{
We also designed the epsilon scheduler that dynamically assigns $ \epsilon $ based on the game score that the agent has achieved; $\epsilon \propto e^{ \frac{ a_{\epsilon} }{ n_{ \text{tst} } } n_{ \text{score} } }$, where $ a_\epsilon $ is the hyper-parameters and $ n_{ \text{tst} } $ is the average testing game score. During training, higher $ n_{ \text{score} } $ exponentially increases $ \epsilon $ while $ a_\epsilon $ controls the slope of the exponential function. Higher $ a_\epsilon $ makes the slope more steep. Intuitively, as the agent exploits the well-known states, $ \epsilon $ is small, encouraging the agent to follow its own policy, and as the agent reaches the under-explored states (i.e., similar to test condition), $ \epsilon $ increases to encourage more diversely. The $ \epsilon $ is normalized and scaled.
The example plot is shown in FIgure \ref{fig:scr2prob}.
}

\input{figure/scr2prob.tex}

{
We conducted a set of ablations with dynamic $ \epsilon $ value in \gdetective, \gpentari, \greverb, \gzorko and \gzorkt.
We used $ \epsilon_{ \text{min} } = \{ 0.0, 0.3 \} $, $ a_{\epsilon} = \{ 3, 9 \} $ and $ \epsilon_{ \text{max} } = \{ 0.7, 1.0 \} $, so total 8 different hyper-parameters.
Figure \ref{fig:tac-dynamic-eps-1} shows fixed $ \epsilon_{ \text{min} } = 0.0 $ with varying $ a_{\epsilon} $ and $ \epsilon_{ \text{max} } $ and Figure \ref{fig:tac-dynamic-eps-1} shows fixed $ \epsilon_{ \text{min} } = 0.3 $.
Other than \gzorkt, TAC with dynamic $ \epsilon $ matches or underperforms TAC with fixed $ \epsilon = 0.3 $.
There are two interesting phenomenons.
(i) Too high $ \epsilon_{ \text{max} } $ results in more unstable learning and lower performance.
This becomes very obvious in \gpentari, \greverb and \gzorko, where regardless of $ \epsilon_{ \text{min} } $ and $ a_{ \epsilon } $, if $ \epsilon_{ \text{max} } = 1.0 $, the learning curve is relatively low.
In \gdetective of Figure \ref{fig:tac-dynamic-eps-1}, the learning becomes much more unstable with $ \epsilon_{ \text{max} } = 1.0 $.
This indicates that even under-explored states, exploitation may still be required.
(ii) Too low $ \epsilon_{ \text{min} } $ results in more unstable learning and lower performance.
Although \gpentari benefits from $ \epsilon_{ \text{min} } = 0.0 $, the learning curves in Figure \ref{fig:tac-dynamic-eps-1} is generally lower and unstable than Figure \ref{fig:tac-dynamic-eps-2}.
This appears to be that despite how much the agent learned the environment, it still needs to act stochastically to collect diverse experiences.


%
}

\section{Limitations}
{
Similar to CALM-DRRN \cite{calm-drrn-trl-lm-jericho-offline-training}, KG-A2C \cite{kg-a2c-trl-jericho} and KG-A2C variants \citep{q-bert-trl-jericho, sha-kg-trl-jericho, hex-rl-im-trl-jericho} that use admissible actions, our method still utilizes admissible actions.
This makes our TAC not suitable for environments that do not provide admissible action set.
In the absence of admissible actions, our TAC requires some prior knowledge of a compact set of more probable actions from LMs or other sources.
This applies to other problems, for instance, applying our proposed method to language-grounded robots requires action candidates appropriate per state that they must be able to sample during training.
The algorithm proposed by \citet{jericho-drrn-tdqn-trl-env} extracts admissible actions by simulating thousands of actions per every step in TGs.
This can be used to extract a compact set of actions in other problems, but it would not be feasible to apply if running a simulation is computationally expensive or risky (incorrect action in real-world robot may result in catastrophic outcomes, such as breakdown).
}

\section{Ethical Considerations}
{
Our proposal may impact other language-based autonomous agents, such as dialogue systems or language-grounded robots.
In a broader aspect, it contributes to the automated decision making, which can be used in corporation and government.
When designing such system, it is important to bring morals and remove bias to be used as intended.



}

\newpage

\onecolumn

\section{\gzorko Gameplay}
\input{data/zork1}

\end{document}

%% file: section/sec-0-abstract.tex
\begin{abstract}
Text-based games (TGs) are language-based interactive environments for reinforcement learning. While language models (LMs) and knowledge graphs (KGs) are commonly used for handling large action space in TGs, it is unclear whether these techniques are necessary or overused. In this paper, we revisit the challenge of exploring the action space in TGs and propose $ \epsilon$-admissible exploration, a minimal approach of utilizing admissible actions, for training phase. Additionally, we present a text-based actor-critic (TAC) agent that produces textual commands for game, solely from game observations, without requiring any KG or LM. Our method, on average across 10 games from Jericho, outperforms strong baselines and state-of-the-art agents that use LM and KG. Our approach highlights that a much lighter model design, with a fresh perspective on utilizing the information within the environments, suffices for an effective exploration of exponentially large action spaces.~\footnote{The code is available at \url{https://github.com/ktr0921/tac}}


%
%
%
\end{abstract}

%% file: section/sec-1-introduction.tex
\section{Introduction} \label{sec:introduction}

{
An intelligent agent that communicates in natural language space has been a long goal of artificial intelligence \cite{fang-etal-2017-learning}.
Text-based games (TGs) best suit this goal, since they allow the agent to \textit{read the textual description of the world} and \textit{write the textual command to the world} \cite{jericho-drrn-tdqn-trl-env, textworld-trl-env}.
In TGs, the agent should perform natural language understanding~(NLU), sequential reasoning and natural language generation~(NLG) to generate a series of actions to accomplish the goal of the game, i.e. adventure or puzzle~\cite{jericho-drrn-tdqn-trl-env}.
The language perspective of TGs foists environments partially observable and action space combinatorially large, making the task challenging. Since TGs alert the player how much the game has proceeded with the game score, reinforcement learning (RL) naturally lends itself as a suitable framework.
%
%
}


{
Due to its language action space, an RL agent in TGs typically deals with a combinatorially large action space, motiving various design choices to account for it. As two seminal works in this space, 
%
%
%
\citet{calm-drrn-trl-lm-jericho-offline-training} trained a language model (LM) to produce admissible actions\footnote{Admissible actions are grounded actions that are guaranteed to change the world state produced by the environment \citep{jericho-drrn-tdqn-trl-env, textworld-trl-env}.} for the given textual observation
%
and then used, under the predicted action list, Deep Reinforcement Relevance Network to estimate the Q value.
%
As an alternative, \citet{kg-a2c-trl-jericho} constructs a knowledge graph~(KG) to prune down action space 
while learning the policy distribution through actor-critic~(AC) method and supervision signal from the admissible actions.
Both paradigms leverage admissible actions at different stages at the cost of imposing additional modules and increasing model complexity.



%

%
%
}

{
%
%
%
%
%
}

{
In this paper, we take a fresh perspective on leveraging the information available in the TG environment to explore the action space without relying on LMs or KGs. 
%
%
We propose a minimal form of utilizing admissibility of actions to constrain the action space during training while allowing the agent to act independently to access the admissible actions during testing.
%
%
More concretely, our proposed training strategy, $ \epsilon $-admissible exploration, leverages the admissible actions via random sampling during training to acquire diverse and useful data from the environment.
{Then, our developed text-based actor-critic (TAC) agent learns the policy distribution without any action space constraints}.
It is noteworthy that our much lighter proposal is under the same condition as other aforementioned methods since all the prior works use admissible actions in training the LM or the agent.
}

{
Our empirical findings, in Jericho, illustrate that TAC with $ \epsilon $-admissible exploration has better or on-par performance in comparison with the state-of-the-art agents that use an LM or KG.
{Through experiments, we observed that while previous methods have their action selections largely dependent on the quality of the LM or KG, sampling admissible actions helps with the action selection and  results in acquiring diverse experiences during exploration.}
While showing a significant success on TGs, we hope our approach encourages alternative perspectives on leveraging action admissibility in other domains of applications where the action space is discrete and combinatorially large.

%
%
%
%
}

%% file: section/sec-2-related-work.tex
\section{Basic Definitions}\label{sec:background}

\stitle{Text-based Games.}
{
TGs are game simulation environments that take natural language commands and return textual description of the world.
They have received significant attention in both NLP  and RL communities in recent years.
\citet{textworld-trl-env} introduced TextWorld, a TG framework that automatically generates textual observation through knowledge base in a game engine.
It has several hyper-parameters to control the variety and difficulty of the game.
\citet{jericho-drrn-tdqn-trl-env} released Jericho, an open-sourced interface for human-made TGs, which has become the de-facto testbed for developments in TG.
}

{
\stitle{Admissible Action.} A list of natural language actions that are guaranteed to be understood by the game engine and change the environment in TGs are called Admissible Actions. The term was introduced in TextWorld while a similar concept also exists in Jericho under a different name, valid actions. \citet{jericho-drrn-tdqn-trl-env} proposed an algorithm that detects a set of admissible actions provided by Jericho suite by constructing a set of natural language actions from every template with detectable objects for a given observation and running them through the game engine to return those actions that changed the world object tree.
}

\stitle{Template-based Action Space.}
{
Natural language actions are built with template ($ \mathbb{T} $) and object ($ \mathbb{O} $) from template-based action space. Each template takes at most two objects.
For instance, a template-object pair (\texttt{take OBJ from OBJ}, \texttt{egg}, \texttt{fridge}) produces a natural language action \texttt{take egg from fridge} while (\texttt{west},\texttt{-},\texttt{-}) produces \texttt{west}.
}

\stitle{Partially Observable Markov Decision Process.}
TG environments can be formalized as Partially Observable Markov Decision Processes (POMDPs).
A POMDP is defined as a 7-tuple, $ ( \mathcal{S} , \mathcal{A} , \mathcal{P} , \mathcal{O} , \mathcal{P}_{o} , \mathcal{R} , \gamma ) $, where $ \mathcal{S} $ and $ \mathcal{A} $ are a set of state and action, and $ \mathcal{P} $ is the state transition probability that maps state-action pair to the next state, $ \Pr ( s_{t+1} | s_{t}, a_{t} ) $.
$ \mathcal{O} $ is a set of observation that depends on the current state via an emission probability, $ \mathcal{P}_{o} \equiv \Pr ( o_{t} | s_{t} ) $.
$ \mathcal{R} $ is an immediate reward signal held between the state and the next state, $ r ( s_{t}, s_{t+1} ) $, and $ \gamma $ is the discount factor. 
The action selection rule is referred to as the policy $ \pi ( a | o ) $, in which the optimal policy acquires the maximum rewards in the shortest move.


\stitle{TG Environment as POMDP.}
{
Three textual observations are acquired from the engine, game feedback $ o_{\text{game}} $, room description $ o_{\text{look}} $, and inventory description $ o_{\text{inv}} $. The game feedback is dependent on the previous action, $ \Pr ( o_{\text{game}, t} | s_{t}, a_{t-1} ) $, while room and inventory descriptions are not, $ \Pr ( o_{\text{look}, t} | s_{t} ) $ and $ \Pr ( o_{\text{inv}, t} | s_{t} ) $. Inadmissible actions do not influence the world state, room and inventory descriptions but change the game feedback changes. Each action is sampled sequentially from template-based action space. For template, we directly sample from observation $ \pi (a_{\mathbb{T}} | o) $ while an object policy is sequentially produced, $ \pi (a_{\mathbb{O}} | o, \hat{a}) $, where $ \hat{a} $ is previously sampled template-object pair. The agent ought to find the optimal policy that maximizes the expected discounted sum of rewards, or the return, $ R_{t} = \sum_{k=0}^{\infty} \gamma^{k} r_{t+k+1} $.
}

\stitle{Traditional Reinforcement Learning.}
{
There are three traditional algorithms in RL, Q-learning (QL), policy gradient (PG) and actor-critic (AC).
QL estimates the return for a given state-action pair, or Q value, $ Q ( s_{t} , a_{t} ) = \mathbb{E} [ \sum_{k=0}^{\infty} \gamma^{k} r_{t+k+1} | s_{t} , a_{t} ] $, then selects the action of the highest Q value.
However, this requires the action space to be countably finite. To remedy this, PG directly learns the policy distribution from the environment such that it maximizes the total return through Monte-Carlo (MC) sampling.
AC combines QL and PG, where it removes MC in PG and updates the parameters per each step with estimated Q value using QL.
This eliminates the high variance of MC as an exchange of a relatively small bias from QL.
}

\section{Related Work on TG Agents in RL}
\label{sec:related-work}
%

We provide a brief overview of widely known TG agents relevant to the work presented in this paper. We empirically compare these in the Section~\ref{sec:experiments:main-results}. 
%
%
%
%




\stitle{Contextual Action LM (CALM)-DRRN} \citep{calm-drrn-trl-lm-jericho-offline-training}
{
uses an LM (CALM) to produce a set of actions for a given textual observation from the TGs.
It is trained to map a set of textual observations to the admissible actions through causal language modeling.
Then, Deep Reinforcement Relevance Network~(DRRN) agent was trained on the action candidates from CALM.
DRRN follows QL, estimating the Q value per observation-action pair.
As a result, CALM removes the need for the ground truth while training DRRN.\footnote{It is noteworthy, orthogonal to the focus of our work, the recently proposed eXploit-Then-eXplore \citep{xtx-mixture-policy-trl-jericho-gpt2-exploitation} uses LM and admissible actions to resolve another challenge, exploration-exploitation dilemma in TGs.}
}



\stitle{Knowledge Graph Advantage Actor Critic (KG-A2C)} \citep{kg-a2c-trl-jericho}
{
uses the AC method
to sequentially sample templates and objects, and KGs for long-term memory and action pruning.
Throughout the gameplay, KG-A2C organizes knowledge triples from textual observation using Stanford OpenIE \citep{openie} to construct a KG.
Then, the KG is used to build state representation along with encoded game observations and constrain object space with only the entities that the agent can reach within KG, i.e. immediate neighbours.
%
They used admissible actions in the cross entropy supervised loss.
}

\stitle{KG-A2C Inspired Agents.}
{
\citet{sha-kg-trl-jericho} proposed SHA-KG that uses stacked hierarchical attention on KG.
Graph attention network (GAT) was applied to sample sub-graphs of KG to enrich the state representation on top of KG-A2C. 
\citet{q-bert-trl-jericho} used  techniques inspired by Question Answering (QA) with LM to construct the KG.
They introduced Q*BERT which uses ALBERT~\citep{albert-lm} fine-tuned on a dataset specific to TGs to perform QA and extract information from textual observations of the game, i.e. ``Where is my current location?".
This improved the quality of KG, and therefore, constituted better state representation.
%
%
%
\citet{ryu-etal-2022-fire} proposed an exploration technique that injects commonsense directly into action selection.
They used log-likelihood score from commonsense transformer \citep{comet-lm-cskg} to re-rank actions.
\citet{hex-rl-im-trl-jericho} investigated explainable generative agent (HEX-RL) and applied hierarchical graph attention to symbolic KG-based state representations.
This was to leverage the graph representation based on its significance in action selection.
They also employed intrinsic reward signal towards the expansion of KG to motivate the agent for exploration (HEX-RL-IM) \citep{hex-rl-im-trl-jericho}.
%
%
}

{
All the aforementioned methods utilize admissible actions in training the LM or agent.
Our proposed method, introduced shortly (\S\ref{sec:tac}), uses admissible actions as action constraints during training without relying on KG or LM.
%
}

%% file: section/sec-4-text-based-actor-critic.tex
\section{Text-based Actor Critic (TAC)} \label{sec:tac}
{
Our agent, Text-based Actor Critic~(TAC), follows the Actor-Critic method with template-object decoder.
We provide an overview of the system in Figure~\ref{fig:tac} and a detailed description in below. We follow the notation introduced earlier in Section~\ref{sec:background}.
}

\stitle{Encoder.}
{
Our design consists of text and state encoders.
Text encoder is a single shared bi-directional GRU with different initial hidden state for different input text, ($ o_{\text{game}} , o_{\text{look}} , o_{\text{inv}} , a_{N} $).
The state representation only takes encoded textual observations while the natural language action $ a_{N} $ is encoded to be used by the critic (introduced shortly).
%
%
State encoder embeds game scores into a high dimensional vector and adds it to the encoded observation.
%
%
This is then, passed through a feed-forward neural network, mapping an instance of observation to state representation without the history of the past information.
}

\stitle{Actor.}
{
The Actor-Critic design is used for our RL component. We describe our generative actor first.
%
Our actor network maps from state representation to action representation.
Then, the action representation is decoded by GRU-based template and object decoders \cite{kg-a2c-trl-jericho}.
Template decoder takes action representation and produces the template distribution and the context vector.
Object decoder takes action representation, semi-completed natural language action and the context from template decoder to produce object distribution sequentially.
%
}
\input{figure/tac.tex}

\stitle{Critic.}
{
Similar to \citep{sac-rl-robotics}, we employed two types of critics for practical purpose, state critic for state value function and state-action critic for state-action value function.
%
Both critics take the state representation as input, but state-action critic takes encoded natural language action as an additional input.
The textual command produced by the decoder is encoded with text encoder and is passed through state-action critic to predict state-action value, or Q value, for a given command.
A more detailed diagram for Actor and Critic is in Appendix D. 
%
%
%
To smooth the training, we introduced target state critic as an exponentially moving average of state critic \citep{ddqn-rl-atari}.
Also, the two state-action critics
 are independently updated to mitigate positive bias in the policy improvement \cite{two-critics-rl}.
We used the minimum of the two enhanced critic networks outputs as our estimated state-action value function.
%
}

\stitle{Objective Function.}
{
Our objective functions are largely divided into two, RL and SL.
RL objectives are for reward maximization $ \mathcal{L}_{ \text{R} } $, state value prediction $ \mathcal{L}_{ \text{V} } $, and state-action value prediction $ \mathcal{L}_{ \text{Q} } $.
We overload the notation of $ \theta $: for instance, $ V_{ \theta } ( o ) $ signifies parameters from the encoder to the critic,  and $ \pi_{ \theta } ( a | o ) $ from the encoder to the actor.
Reward maximization is done as follows,
{
\setlength{\abovedisplayskip}{3pt}
\setlength{\belowdisplayskip}{3pt}
\begin{gather}
    \mathcal{L}_{ \text{R} } = - \mathbb{E} \left[ A ( o , a ) \nabla_{ \theta } \ln \pi_{ \theta } \left( a | o \right) \right],
    \\
    A ( o , a ) = Q_{ \theta } ( o , a ) - V_{ \theta } ( o ),
\end{gather}
}%
where $ A ( o , a ) $ is the normalized advantage function with no gradient flow.
{\small
\setlength{\abovedisplayskip}{3pt}
\setlength{\belowdisplayskip}{3pt}
\begin{gather}
    \mathcal{L}_{ \text{V} } = \mathbb{E} \left[ \nabla_{ \theta } \left( V_{ \theta } ( o ) -  \left( r + \gamma V_{ \bar{ \theta } } ( o' ) \right) \right) \right],
    \\
    \mathcal{L}_{ \text{Q} } = \mathbb{E} \left[ \nabla_{ \theta } \left( Q_{ \theta } ( o , a ) -  \left( r + \gamma V_{ \bar{ \theta } } ( o' ) \right) \right) \right],
\end{gather}
}%
where $ o' $ is observation in the next time step and $ \bar{\theta} $ signifies the parameters containing the target state critic, updated as moving average with $ \tau $,
{
\setlength{\abovedisplayskip}{3pt}
\setlength{\belowdisplayskip}{3pt}
\begin{equation}
    \bar{\theta}_{v} = \tau \theta_{v} + (1 - \tau) \bar{\theta}_{v}.
\end{equation}
}%
}
{
Our SL updates the networks to produce valid templates and valid objects,
{
\setlength{\abovedisplayskip}{3pt}
\setlength{\belowdisplayskip}{3pt}
\begin{gather}
    \begin{aligned}
        \mathcal{L}_{ \mathbb{T} } = & \frac{1}{ | \mathbb{T} | } \sum_{ a_{ \mathbb{T} } \in \mathbb{T} } 
        (
        y_{ a_{ \mathbb{T} } } \ln \left( \pi_{ \theta } ( a_{ \mathbb{T} } | o ) \right)
        \\
        & + ( 1 - y_{ a_{ \mathbb{T} } } ) \left( 1 - \ln \left( \pi_{ \theta } ( a_{ \mathbb{T} } | o ) \right) \right)
        ),
    \end{aligned}
    \\
    \begin{aligned}
        \mathcal{L}_{ \mathbb{O} } = & \frac{1}{ | \mathbb{O} | } \sum_{ a_{ \mathbb{O} } \in \mathbb{O} } 
        (
        y_{ a_{ \mathbb{O} } } \ln \left( \pi_{ \theta } ( a_{ \mathbb{O} } | o , \hat{a} ) \right)
        \\
        & + ( 1 - y_{ a_{ \mathbb{O} } } ) \left( 1 - \ln \left( \pi_{ \theta } ( a_{ \mathbb{O} } | o , \hat{a} ) \right) \right)
        ),
        \label{eq:sup_obj}
    \end{aligned}
    \\
    y_{ a_{ \mathbb{T} } } =
    \begin{cases}
    1 & a_{ \mathbb{T} } \in \mathbb{T}_{a} \\
    0 & \text{otherwise}
    \end{cases}
    \quad
    y_{ a_{ \mathbb{O} } } =
    \begin{cases}\nonumber
    1 & a_{ \mathbb{O} } \in \mathbb{O}_{a} \\
    0 & \text{otherwise}
    \end{cases}
\end{gather}
}%
where $ \mathcal{L}_{ \mathbb{T} } $ and $ \mathcal{L}_{ \mathbb{O} } $ are the cross entropy losses over the templates ($\mathbb{T}$) and objects ($\mathbb{O}$).
Template and object are defined as $ a_{ \mathbb{T} } $ and $ a_{ \mathbb{O} } $, while $ \hat{a} $ is the action constructed by previously sampled template and object.
Positive samples, $ y_{a_{ \mathbb{T} }} $ and $ y_{a_{ \mathbb{O} }} $, are only if the corresponding template or object are in the admissible template ($ \mathbb{T}_{a} $) or admissible object ($ \mathbb{O}_{a} $).\footnote{Eq. \ref{eq:sup_obj} is calculated separately for two objects in a single template, where the admissible object space ($ \mathbb{O}_{a} $) is conditioned on the previously sampled template and object.}
The final loss function is constructed with $\lambda$ coefficients to control for trade-offs,
{
\setlength{\abovedisplayskip}{3pt}
\setlength{\belowdisplayskip}{3pt}
\begin{equation}
    \mathcal{L} = 
    \lambda_{ \text{R} } \mathcal{L}_{ \text{R} }
    + \lambda_{ \text{V} } \mathcal{L}_{ \text{V} }
    + \lambda_{ \text{Q} } \mathcal{L}_{ \text{Q} }
    + \lambda_{ \mathbb{T} } \mathcal{L}_{ \mathbb{T} }
    + \lambda_{ \mathbb{O} }  \mathcal{L}_{ \mathbb{O}. }
\end{equation}
}
}

\begin{table*}[t!]
\centering
\footnotesize
\scalebox{0.98}{
\begin{tabular}{lccccccc}
\toprule
& \textsc{LM-based} & \multicolumn{5}{c}{\textsc{KG-based}}                        & \\ 
\cmidrule(lr){2-2}\cmidrule(lr){3-7}
                          Games & CALM-DRRN        & KG-A2C        & SHA-KG & Q*BERT        & HEX-RL           & HEX-RL-IM & TAC                        \\ 
\cmidrule(lr){1-1}\cmidrule(lr){2-2}\cmidrule(lr){3-7}\cmidrule(lr){8-8}
\gbalances                 & 9.1               & \textbf{10.0} & 9.8    & \textbf{10.0} & \textbf{10.0}    & \textbf{10.0}    & \textbf{10.0 $ \pm $ 0.1}  \\
\gdeephome                & 1.0               & 1.0           & 1.0    & 1.0           & 1.0              & 1.0              & \textbf{25.4 $ \pm $ 3.2}  \\
\gdetective               & \textbf{289.7}    & 207.9         & 246.1  & 274.0         & 276.7            & 276.9            & 272.3 $ \pm $ 23.3         \\
\glibrary                 & 9.0               & 14.3          & 10.0   & \textbf{18.0} & 15.9             & 13.8             & \textbf{18.0 $ \pm $ 1.2}  \\
\gludicorp               & 10.1              & 17.8          & 17.6   & \textbf{18.0} & 14.0             & 17.6             & 7.7 $ \pm $ 2.5            \\
\gpentari                & 0.0               & 50.7          & 48.2   & 50.0          & 34.6             & 44.7             & \textbf{53.2 $ \pm $ 2.9}  \\
\gtemple                  & 0.0               & 7.6           & 7.9    & \textbf{8.0}  & \textbf{8.0}     & \textbf{8.0}     & 5.8 $ \pm $ 2.3            \\
\gzorko                  & 30.4              & 34.0          & 33.6   & 35.0          & 29.8             & 30.2             & \textbf{46.3 $ \pm $ 5.0}  \\
\gzorkt                  & 0.5               & 0.1           & 0.7    & 0.1           & $-$              & $-$              & \textbf{1.6 $ \pm $ 1.2}  \\
\gztuu                  & 3.7               & 5.0           & 5.0    & 5.0           & 5.0              & 5.1              & \textbf{33.2 $ \pm $ 26.3} \\ \midrule
\textsc{Normalized Mean}                      & 0.1549            & 0.2475        & 0.2490 & 0.2788        & 0.2722$^\dagger$ & 0.2834$^\dagger$ & \textbf{0.3307}            \\ \bottomrule
\end{tabular}
}
\caption{Game score comparison over 10 popular game environments in Jericho, with best results highlighted by \textbf{boldface}. We only included algorithms that reported the end performance. $ ^\dagger $HEX-RL and HEX-RL-IM did not report the performance in \gzorkt and are not open-sourced, so the mean average did not account \gzorkt.}
\label{tab:main}
\end{table*}

\input{figure/tac-main-5}

Our algorithm is akin to vanilla A2C proposed by \citet{kg-a2c-trl-jericho} with some changes under our observations. A detailed comparison and qualitative analysis are in Appendix E and F.

\paragraph{$ \epsilon $-admissible Exploration.}
{
We use a simple exploration technique during training, which samples the next action from admissible actions with $ \epsilon $ probability threshold.
For a given state $ s $, define $ \mathcal{A}_{a} ( s ) \subseteq \mathcal{A}_{N} $ as an admissible action subset of all natural language actions set. We sample an action directly from admissible action set under uniform distribution, $ a_{N} \sim \mathcal{U} ( \mathcal{A}_{a} ( s ) ) $.
Formally, we uniformly sample $ p \in [0, 1] $ per every step,
{
\setlength{\abovedisplayskip}{3pt}
\setlength{\belowdisplayskip}{3pt}
\begin{equation}
    \beta ( a | s )
    = \begin{cases}
        \mathcal{U} ( \mathcal{A}_{a} ( s ) ) & p < \epsilon \\
        \pi ( a | s ) & p \geq \epsilon \\
    \end{cases}
\end{equation}
}%
This collects diverse experiences from altering the world with admissible actions. We also tried a variant where the $\epsilon$ is selected adaptively given the game score the agent has achieved. However, this variant under-performed the static $\epsilon$. See Appendix I for more details on this and the results.

%% file: figure/tac.tex
\begin{figure*}[ht!]
    \centering
    \includegraphics[width=0.9\textwidth]{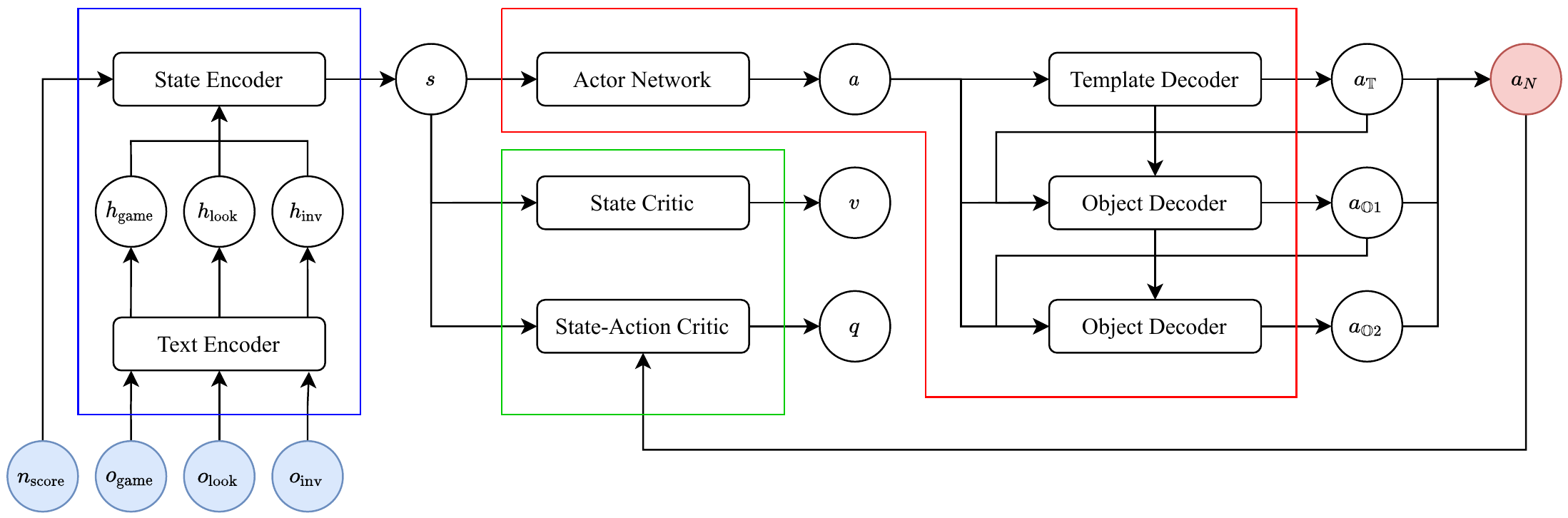}
    \caption{Text-based Actor-Critic (TAC); A blue circle is the input to the encoder, ($ n_{ \text{score} }, o_{ \text{game} }, o_{ \text{look} }, o_{ \text{inv} } $) representing (game score, game feedback, room description, inventory), while a red circle is the output from actor, $ a_{ N } $ representing natural language action. Blue, red and green boxes indicate encoder, actor and critic, respectively.}
    \label{fig:tac}
    \vspace{-2mm}
\end{figure*}

%% file: figure/tac-main-5.tex
\begin{figure*}[t!]
    \centering
    \includegraphics[width=1\textwidth]{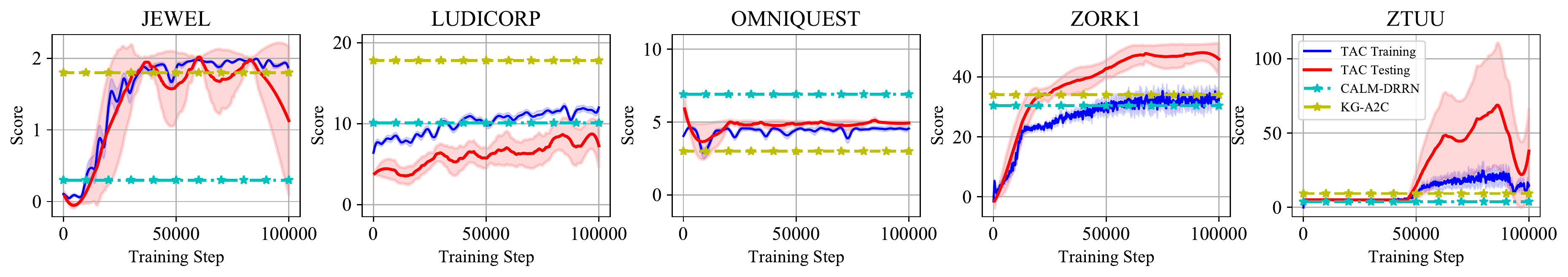}
    \caption{The full learning curve of TAC on five games in Jericho suite. Blue and red plots are training and testing game score while cyan and yellow star marker line signify CALM-DRRN and KG-A2C.}
    \label{fig:tac-main-5}
    \vspace{-2mm}
\end{figure*}

%% file: section/sec-5-experiments.tex
\section{Experiments} \label{sec:experiments}

%
%
%
%
%
%
%
%
%
%

{
In this section, we provide a description of our experimental details and discuss the results.
We selected a wide variety of agents~(introduced in Section~\ref{sec:related-work}) utilizing the LM or the KG:
CALM-DRRN \citep{calm-drrn-trl-lm-jericho-offline-training} and KG-A2C \citep{kg-a2c-trl-jericho} as baselines, and SHA-KG \citep{sha-kg-trl-jericho}, Q*BERT \citep{q-bert-trl-jericho}, HEX-RL and HEX-RL-IM \citep{hex-rl-im-trl-jericho} as state-of-the-art (SotA).
%
}

\stitle{Experimental Setup.} \label{sec:experiments:setup}
{
Similar to KG-A2C, we train our agent on 32 parallel environments with 5 random seeds. We trained TAC on games of Jericho suite with 100k steps and evaluated with 10 episodes per every 500 training step. During the training, TAC uses uniformly sampled admissible action for a probability of $ \epsilon $ and during the testing, it follows its policy distribution generated from the game observations. We used prioritized experience replay (PER) as our replay buffer \citep{per}. We first fine-tune TAC on \gzorko, then apply the same hyper-parameters for all the games. The details of our hyper-parameters can be found in Appendix A.
Our final score is computed as the average of 30 episodic testing game scores.  Additionally, our model has a parameter size of less than 2M, allowing us to run the majority of our experiments on CPU (Intel Xeon Gold 6150 2.70 GHz).
The full parameter size in \gzorko and the training time comparison can be found in Appendices B and C.
}

\input{figure/tac-ablation}

\input{figure/tac-regularization-3.tex}

\input{section/sec-5-1-main}

\input{section/sec-5-2-ablation}

%% file: figure/tac-ablation.tex
\begin{figure*}[t!]
    \centering
    \includegraphics[width=1\textwidth]{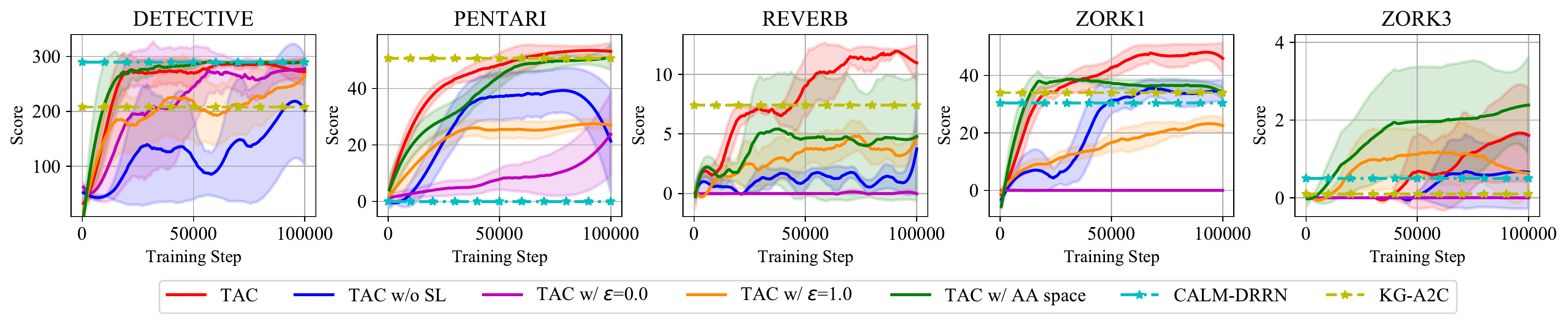}
    \caption{Ablation study on five popular games in Jericho suite. Four different ablation are conducted with SL, $ \epsilon = 0.0 $, $ \epsilon = 1.0 $, and with full admissible constraints during training (Admissible Action space). Similar to the previous figure, CALM-DRRN and KG-A2C are added for comparison.}
    \label{fig:tac-ablation}
\end{figure*}

%% file: figure/tac-regularization-3.tex
\begin{figure*}[t]
    \centering
    \includegraphics[width=0.555\textwidth]{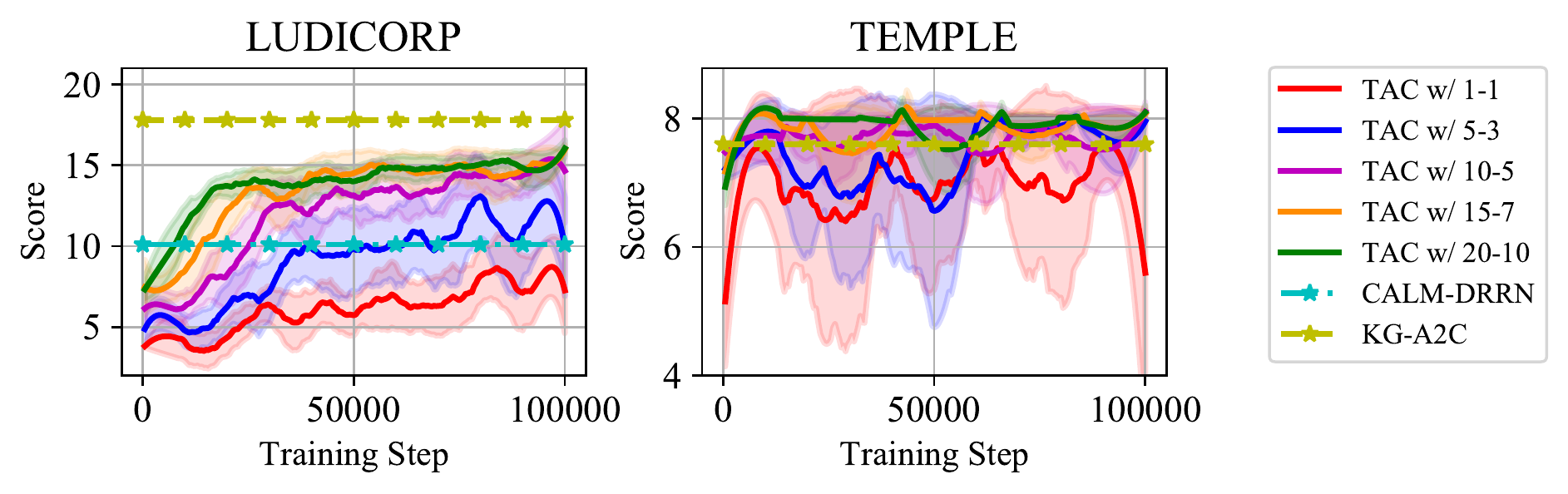}
    \includegraphics[width=0.435\textwidth]{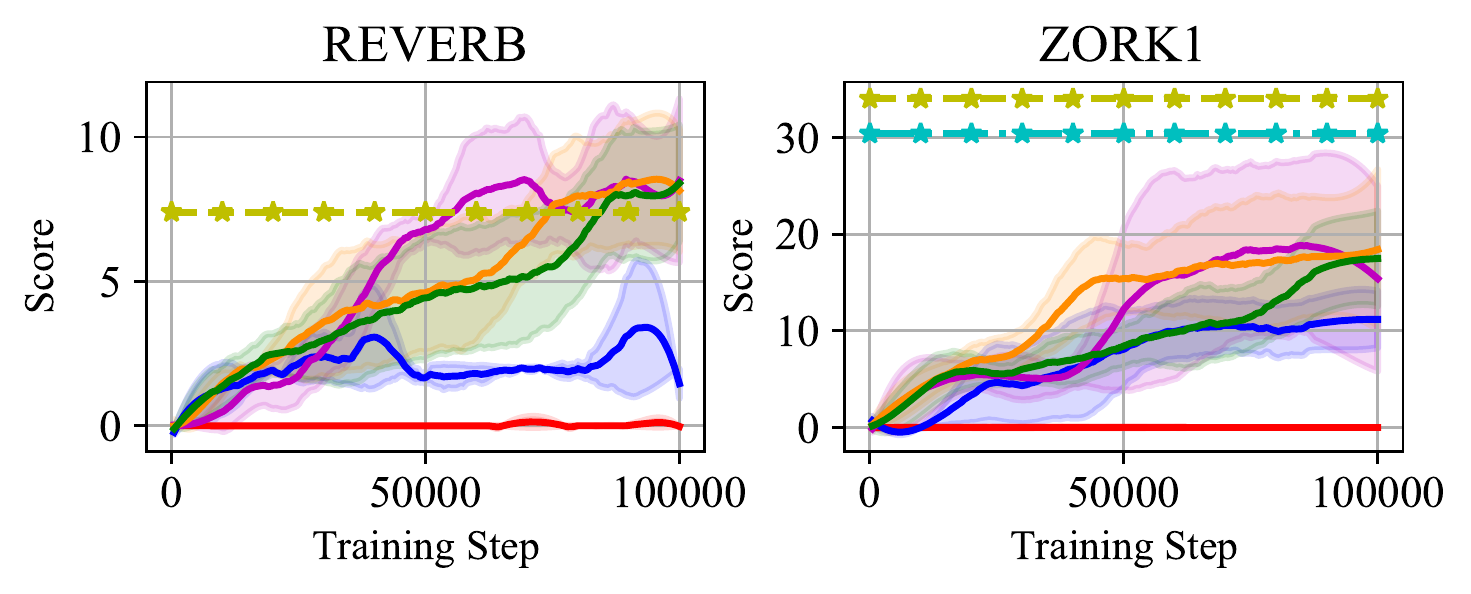}
    \caption{The learning curve of TAC for stronger supervised signals where 5-3 signifies $ \lambda_{ \mathbb{T} } = 5 $ and $ \lambda_{ \mathbb{O} } = 3 $. Left two plots are with $ \epsilon = 0.3 $ and right two are with $ \epsilon =0 $.}
    \vspace{-2mm}
    \label{fig:tac-regularization}
\end{figure*}

%% file: section/sec-5-1-main.tex
\subsection{Main Results} \label{sec:experiments:main-results}
{
Table \ref{tab:main} reports the results for baselines, SotAs and TAC on 10 popular Jericho games.
TAC attains the new SotA scores in 5 games. Apart from \gpentari, TAC surpasses 4 games with a large margin, where all of the other agents fail to pass the performance bottleneck (\gdeephome with 1, \gzorko with 35, \gzorkt with 1, and \gztuu with 5). In \gdetective, TAC matches many SotAs, but falls short in \gludicorp and \gtemple. Nevertheless, TAC achieves the highest mean score over LM or KG-based methods.

{
On a larger set of 29 games in comparison with the baselines, TAC surpasses CALM-DRRN in \textit{14 out of 29 games} and KG-A2C in \textit{16 out of 29 games} and achieves more than $ \sim 50 \% $ higher score than both CALM-DRRN and KG-A2C with normalized mean score.
Per game, in \gsorcerer, \gspirit, \gzorkt and \gztuu, TAC achieves at least $ \sim 200 \% $ and at most $ \sim 400 \% $ higher score..
In \gacorncourt, \gdeephome and \gdragon, both CALM-DRRN and KG-A2C fails to achieve any game score (approximately 0), but TAC achieves the score of $ + 3.4 $, $ + 25.4 $ and $ + 2.81 $
For detailed game scores and the full learning curves on 29 games, please refer to Appendix G.
}

There are a few games that TAC under-performs. We speculate three reasons for this: over-fitting, exploration, and catastrophic forgetting. 
For instance, as illustrated by the learning curves of TAC in Figure \ref{fig:tac-main-5}, \gludicorp appears to acquire more reward signals during training, but fails to achieve them during testing.
We believe this is because the agent is over-fitted to spurious features in specific observations \cite{observational-overfitting}, producing inadmissible actions for a given state that are admissible in other states.
On the other hand, TAC in \gomniquest cannot achieve a game score more than 5 in both training and testing.
This is due to the lack of exploration, where the agent is stuck at certain states because the game score is too far to reach.
This, in fact, occurs in \gzorkt and \gztuu for some random seeds, where few seeds in \gzorkt do not achieve any game score while \gztuu achieves 10 or 13 only, resulting in high variance.
Finally, catastrophic forgetting \cite{catastrophic-forgetting} is a common phenomenon in TGs \cite{jericho-drrn-tdqn-trl-env, kg-a2c-trl-jericho}, and this is also observed in \gjewel with TAC.
%
}

\stitle{Training Score vs. Testing Score.}
{
Figure \ref{fig:tac-main-5} shows that the game scores during training and testing in many games are different.
There are three interpretations for this: (i) the $ \epsilon $-admissible exploration triggers negative rewards since it is uniformly sampling admissible actions. It is often the case that negative reward signal triggers termination of the game, i.e. $ -10 $ score in \gzorko, so this results in episodic score during training below testing. (ii) the $ \epsilon $-admissible exploration sends the agent to the rarely or never visited state, which is commonly seen in \gztuu. This induces the agent taking useless actions that would not result in rewards since it does not know what to do. (iii)  Over-fitting where testing score is lower than training score.  This occurs in \gludicorp, where the agent cannot escape certain states with its policy during testing.
$ \epsilon $-admissible exploration lets the agent escape from these state during training, and therefore, achieves higher game score.
}

{
}

%% file: section/sec-5-2-ablation.tex
\subsection{Ablation}\label{sec:ablation}
{
}

\stitle{$ \epsilon $-Admissible Exploration.}
{
To understand how $ \epsilon $ influences the agent, ablations with two $ \epsilon $ values, $ 0.0 $ and $ 1.0 $, on five selective games were conducted.
%
As shown in Figure \ref{fig:tac-ablation}, in the case of $ \epsilon = 0.0 $, the agent simply cannot acquire reward signals.
TAC achieves 0 game score in \greverb, \gzorko and \gzorkt while it struggles to learn in \gdetective and \gpentari.
This indicates that the absence of $ \epsilon $-admissible exploration results in meaningless explorations until admissible actions are reasonably learned through supervised signals.
%
%
With $ \epsilon = 1.0 $, learning becomes unstable since this is equivalent to no exploitation during training, not capable of observing reward signals that are far from the initial state.
Hence, tuned $ \epsilon $ is important to allow the agent to cover wider range of states (exploration) while acting from its experiences (exploitation).
}

\stitle{Supervised Signals.}
{
%
According to the Figure \ref{fig:tac-ablation}, removing SL negatively affects the game score.
%
This is consistent with the earlier observations \cite{kg-a2c-trl-jericho} reporting that KG-A2C without SL achieves no game score in \gzorko.
However, as we can observe, TAC manages to retain some game score, which could be reflective of  the positive role of $ \epsilon $-admissible exploration, inducing similar behaviour to SL.
%
}

{
From the observation that the absence of SL degrades the performance, we hypothesize that SL induces a regularization effect.
We ran experiments with various strengths of supervised signals by increasing $ \lambda_{ \mathbb{T} } $ and $ \lambda_{ \mathbb{O} } $ in \gludicorp and \gtemple, in which TAC attains higher scores at training compared with testing.
As seen in Figure \ref{fig:tac-regularization} (left two plots), higher $ \lambda_{ \mathbb{T} } $ and $ \lambda_{ \mathbb{O} } $ relaxes over-fitting, reaching the score from 7.7 to 15.8 in \gludicorp and from 5.8 to 8.0 in \gtemple.
Since SL is not directly related to rewards, this supports that SL acts as regularization.
Further experimental results on \gzorko is in Appendix H.
}

{
To further examine the role of admissible actions in SL, we hypothesize that SL is responsible for guiding the agent in the case that the reward signal is not collected.
To verify this, we excluded $ \epsilon $-admissible exploration and ran TAC with different $ \lambda_{ \mathbb{T} } $ and $ \lambda_{ \mathbb{O} } $ in \greverb and \gzorko, in which TAC fails to achieve any score.
According to Figure \ref{fig:tac-regularization} (right two plots), TAC with stronger SL and $ \epsilon = 0.0 $ achieves game scores from 0 to 8.3 in \greverb, and from 0 to 18.3 in \gzorko, which suggests that SL acts as guidance.
However, in the absence of $ \epsilon $-admissible exploration, despite the stronger supervised signals, TAC cannot match the scores using $ \epsilon $-admissible exploration.
}


\stitle{Admissible Action Space During Training.}
{
To examine if constraining the action space to admissible actions during training leads to better utilization, we ran an ablation by masking template and object with admissible actions at training time. This leads to only generating admissible actions.
Our plots in Figure \ref{fig:tac-ablation} show that there is a reduction in the game score in \gpentari, \greverb and \gzorko while \gdetective and \gzorkt observe slight to substantial increases, respectively.
We speculate that the performance decay is due to the exposure bias \cite{exposure-bias-scheduled-sampling} introduced from fully constraining the action space to admissible actions during training.
This means the agent does not learn how to act when it receives observations from inadmissible actions at test phase.
However, for games like \gzorkt, where the agent must navigate through the game to acquire sparse rewards, this technique seems to help.
}

\begin{table}[t]
\centering
\scriptsize
\scalebox{0.92}{
\begin{tabular}{{>{\RaggedRight}p{0.8cm}>{\RaggedRight}p{6.5cm}}}
\toprule
Game & Kitchen. On the table is an elongated brown sack, smelling of hot peppers. A bottle is sitting on the table. The glass bottle contains: A quantity of water.\\
\cdashline{2-2}
Inventory  &You are carrying: A painting,  A brass lantern (providing light)\\
\cdashline{2-2}
Room & Kitchen. You are in the kitchen of the white house. A table seems to have been used recently for the preparation of food. A passage leads to the west and a dark staircase can be seen leading upward. A dark chimney leads down and to the east is a small window which is open. On the table is an elongated brown sack, smelling of hot peppers. A bottle is sitting on the table. The glass bottle contains: A quantity of water\\\midrule
LM Actions & `close bottle', `close door', `down', `drink water', `drop bottle', `drop painting', `east', `empty bottle', `get all', `get bottle', `get on table', `get painting', `get sack', `north', `open bottle', `out', `pour water on sack', `put candle in sack', `put painting in sack', `put painting on sack', `put water in sack', `south', `take all', `take bottle', `take painting', `take sack', `throw painting', `up', `wait', `west'\\ 
\cdashline{2-2}
KG Objects & `a', `all', `antique', `board', `bottle', `brass', `chimney', `dark', `door', `down', `east', `exit', `front',  `grue', `house', `is', `kitchen', `lantern', `large', `light', `narrow', `north', `of', `passage', `path', `quantity', `rug', `south', `staircase', `table', `to', `trap', `trophy', `up', `west', `white', `window', `with'\\ \cdashline{2-2}
Admiss. Actions & \textcolor{red}{`close window'}, `east', \textcolor{red}{`jump'}, `open bottle', \textcolor{brown}{`open sack'}, \textcolor{red}{`put down all'}, \textcolor{red}{`put down light'},  \textcolor{brown}{`put down painting'}, \textcolor{red}{`put light on table'}, \textcolor{red}{`put out light'}, \textcolor{brown}{`put painting on table'}, `take all', `take bottle',  \textcolor{blue}{`take sack'}, \textcolor{red}{`throw light at window'}, `up', `west'\\ 
\bottomrule
\end{tabular}}
\caption{Action space for a game observation (top panel) for CALM (LM), KG-A2C (KG), and the Admissible Action sets.  \textcolor{red}{Red} and \textcolor{blue}{blue} colored actions are the actions missed by either CALM or KG-A2C. \textcolor{brown}{Brown} are the actions missed by both, and blacks are actions covered by both.}
\label{tab:qualitative}
\end{table}

\subsection{Qualitative Analysis}
{
In this section, we show how CALM and KG-A2C restrict their action space.
Table \ref{tab:qualitative} shows a snippet of the gameplay in \gzorko.
Top three rows are the textual observations and the bottom three rows are the actions generated by CALM, the objects extracted from KG in KG-A2C, and the admissible actions from the environment.
CALM produces 30 different actions, but still misses 10 actions out of 17 admissible actions.
Since DRRN learns to estimate Q value over generated 30 actions, those missing admissible actions can never be selected, resulting in a lack of exploration.
On the other hand, KG-generated objects do not include `sack' and `painting', which means that the KG-A2C masks these two objects out from their object space.
Then, the agent neglects any action that includes these two object, which also results in a lack of exploration.
%
}

%% file: section/sec-6-discussion.tex
\section{Discussion} \label{sec:discussion}

\stitle{Supervised Learning Loss.}
{
%
Intuitively, RL is to teach the agent \textit{how to complete the game} while SL is to teach \textit{how to play the game}.
If the agent never acquired any reward signal, learning is only guided by SL.
This is equivalent to applying imitation learning to the agent to follow more probable actions, a.k.a. admissible actions in TGs.
However, in the case where the agent has reward signals to learn from, SL turns into regularization (\S\ref{sec:ablation}), inducing a more uniformly distributed policies.
In this sense, SL could be considered as the means to introduce the effects similar to entropy regularization in \citet{kg-a2c-trl-jericho}.
}


\paragraph{Exploration as Data Collection.}
{
In RL, the algorithm naturally collects and learns from data.
Admissible action prediction from LM is yet to be accurate enough to replace the true admissible actions \citep{modelling-worlds-in-text, calm-drrn-trl-lm-jericho-offline-training}.
This results in poor exploration and the agent may potentially never reach a particular state.
On the other hand, KG-based methods \cite{kg-a2c-trl-jericho, hex-rl-im-trl-jericho, sha-kg-trl-jericho, xu-etal-2021-generalization-text, xu-etal-2022-perceiving,ryu-etal-2022-fire} must learn admissible actions before exploring the environment meaningfully.
This will waste many samples since the agent will attempt inadmissible actions, collecting experiences of the unchanged states.
Additionally, its action selection is largely dependent on the quality of KG.
The missing objects from KG may provoke the same effects as LM, potentially obstructing navigating to a particular state.
In this regards, $ \epsilon $-admissible exploration can overcome the issue by promoting behaviour that the agent would take after learning admissible actions fully.
Under such conditions that a compact list of actions is either provided the environment or extracted by algorithm \citep{jericho-drrn-tdqn-trl-env}, our approach can be employed.
Intuitively, this is similar to playing the game with a game manual but not a ground truth to complete the game, which leads to collecting more meaningful data.
%
It also collects more diverse data due to the stochasticity of exploration.
Hence, TAC with $ \epsilon $-admissible exploration can learn \textit{how to complete the game} with minimal knowledge of \textit{how to play the game}.
}

\paragraph{Bias in Exploration.}
{
Our empirical results from adaptive $\epsilon$ experiments in Appendix I suggest that reasonable $ \epsilon $ is required for both under-explored states and well-explored states.
This could indicate that diverse data collection is necessary regardless of how much the agent knows about the game while $ \epsilon $ value should not be too high such that the agent can exploit.
Finally, from our ablation, fully constraining action space to admissible actions degrades performance. 
This could be a sign of exposure bias, which is a typical issue in NLG tasks \cite{nlg-quantify-exposure-bias, conv-qa-exposure-bias} and occurs between the training-testing discrepancy due to the teacher-forcing done at training \cite{nlg-quantify-exposure-bias}.
In our setting, this phenomena could potentially occur if the agent only learns from admissible actions at training time.
Since $ \epsilon $-admissible exploration allows a collection of experiences of any actions (i.e., potentially inadmissible actions) with probability of $ 1 - \epsilon $, TAC with reasonable $ \epsilon $ learns from high quality and unbiased data.
Our observations indicate that both the algorithm that learns from data, and the exploration to acquire data are equally important.
%
%
}

%% file: section/sec-7-conclusion.tex
\section{Conclusion} \label{sec:conclusion}
{

%
Text-based Games (TGs) offer a unique framework for developing RL agents for goal-driven and contextually-aware natural language generation tasks. In this paper we took a fresh approach in utilizing the information from the TG environment, and in particular the admissibility of actions during the exploration phase of RL agent. We introduced a language-based actor critic method (TAC) with a simple $ \epsilon $-admissible exploration. The core of our algorithm is the utilization of admissible actions in training phase to guide the agent exploration towards collecting more informed experiences. Compared to state-of-the-art approaches with more complex design, our light TAC design achieves substantially higher game scores across 10-29 games. 

{
We provided insights into the role of action admissibility and supervision signals during training and the implications at test phase for an RL agent.
Our analysis showed that supervised signals towards admissible actions act as guideline in the absence of reward signal, while serving a regularization role in the presence of such signal.
We demonstrated that reasonable $ \epsilon $ probability threshold is required for high quality unbiased experience collection during the exploration phase. 
}


%
%

%

%

%
%
%
%
%
}



%% file: figure/tac-decoder.tex
\begin{figure*}[htbp!]
    \centering
    \includegraphics[width=0.85\textwidth]{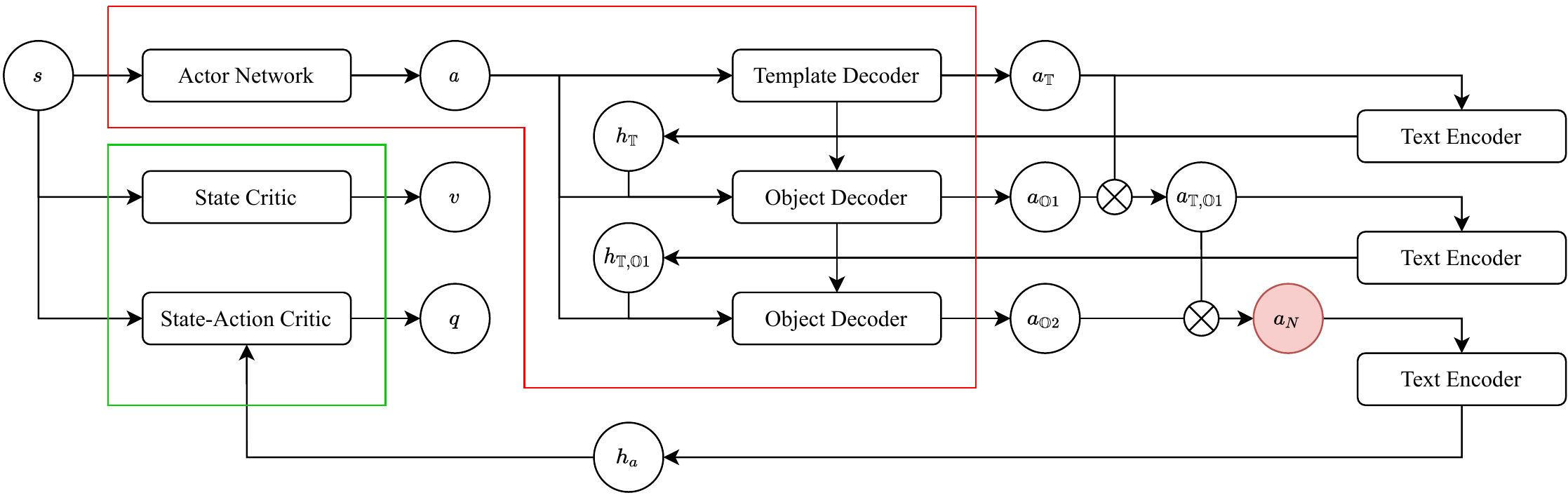}
    \caption{The details of actor and critic of text-based actor-critic; State representation is the input to actor-critic while a red circle is the output from actor, $ a_{ N } $ representing natural language action. Red and green boxes indicate actor and critic, respectively.}
    \label{fig:tac-decoder}
\end{figure*}

%% file: table/tab-hyper-parameters.tex
\begin{table}[ht!]
\centering
\scriptsize
\begin{tabular}{l|c}
\hline
\multicolumn{2}{c}{Training}               \\ \hline
\# of parallel environments  & 32          \\
$ p_{ va } $                 & 0.3         \\ \hline
\multicolumn{2}{c}{Optimization}           \\ \hline
Batch size                   & 64          \\
Learning rate                & $ 10^{-4} $ \\
Weight decay                 & $ 10^{-6} $ \\
Clip                         & 5           \\
$ \gamma $                   & 0.95        \\
$ \tau $                     & 0.001       \\ \hline
\multicolumn{2}{c}{Parameter size}         \\ \hline
Word embedding dimension     & 100         \\
Hidden dimension             & 128         \\ \hline
\multicolumn{2}{c}{Replay buffer}          \\ \hline
Memory size                  & $ 10^{5} $  \\
$ \alpha $                   & 0.7         \\
$ \beta $                    & 0.3         \\ \hline
\multicolumn{2}{c}{Weights for objectives} \\ \hline
$ \lambda_{ \text{R} } $      & 1.0         \\
$ \lambda_{ \text{V} } $      & 1.0         \\
$ \lambda_{ \text{Q} } $      & 1.0         \\
$ \lambda_{ \mathbb{T} } $    & 1.0         \\
$ \lambda_{ \mathbb{O} } $    & 1.0         \\ \hline
\end{tabular}
\caption{Hyper-parameters for main experiments.}
\label{tab:hyper-parameters}
\end{table}

%% file: table/tab-parameter-size-zork1.tex
\begin{table}[ht!]
\centering
\footnotesize
\caption{Parameter size for \gzorko.}
\scalebox{0.6}{
\begin{tabular}{l|c}
\hline
Name                                                & Size           \\ \hline
text\_encoder\_network.embedding.weight             & {[}8000,100{]} \\
text\_encoder\_network.embedding\_sa.weight         & {[}4,128{]}    \\
text\_encoder\_network.encoder.weight\_ih\_l0       & {[}384,100{]}  \\
text\_encoder\_network.encoder.weight\_hh\_l0       & {[}384,128{]}  \\
text\_encoder\_network.encoder.bias\_ih\_l0         & {[}384{]}      \\
text\_encoder\_network.encoder.bias\_hh\_l0         & {[}384{]}      \\
state\_network.embedding\_score.weight              & {[}1024,128{]} \\
state\_network.tf.weight                            & {[}128,384{]}  \\
state\_network.tf.bias                              & {[}128{]}      \\
state\_network.fc1.weight                           & {[}128,128{]}  \\
state\_network.fc1.bias                             & {[}128{]}      \\
state\_network.fc2.weight                           & {[}128,128{]}  \\
state\_network.fc2.bias                             & {[}128{]}      \\
state\_network.fc3.weight                           & {[}128,128{]}  \\
state\_network.fc3.bias                             & {[}128{]}      \\
state\_network.s.weight                             & {[}128,128{]}  \\
state\_network.s.bias                               & {[}128{]}      \\
state\_critic.fc1.weight                            & {[}128,128{]}  \\
state\_critic.fc1.bias                              & {[}128{]}      \\
state\_critic.fc2.weight                            & {[}128,128{]}  \\
state\_critic.fc2.bias                              & {[}128{]}      \\
state\_critic.fc3.weight                            & {[}128,128{]}  \\
state\_critic.fc3.bias                              & {[}128{]}      \\
state\_critic.v.weight                              & {[}1,128{]}    \\
state\_critic.v.bias                                & {[}1{]}        \\
actor\_network.fc1.weight                           & {[}128,128{]}  \\
actor\_network.fc1.bias                             & {[}128{]}      \\
actor\_network.fc2.weight                           & {[}128,128{]}  \\
actor\_network.fc2.bias                             & {[}128{]}      \\
actor\_network.fc3.weight                           & {[}128,128{]}  \\
actor\_network.fc3.bias                             & {[}128{]}      \\
actor\_network.a.weight                             & {[}128,128{]}  \\
actor\_network.a.bias                               & {[}128{]}      \\
state\_action\_critic\_1.fc1.weight                 & {[}128,256{]}  \\
state\_action\_critic\_1.fc1.bias                   & {[}128{]}      \\
state\_action\_critic\_1.fc2.weight                 & {[}128,128{]}  \\
state\_action\_critic\_1.fc2.bias                   & {[}128{]}      \\
state\_action\_critic\_1.fc3.weight                 & {[}128,128{]}  \\
state\_action\_critic\_1.fc3.bias                   & {[}128{]}      \\
state\_action\_critic\_1.q.weight                   & {[}1,128{]}    \\
state\_action\_critic\_1.q.bias                     & {[}1{]}        \\
state\_action\_critic\_2.fc1.weight                 & {[}128,256{]}  \\
state\_action\_critic\_2.fc1.bias                   & {[}128{]}      \\
state\_action\_critic\_2.fc2.weight                 & {[}128,128{]}  \\
state\_action\_critic\_2.fc2.bias                   & {[}128{]}      \\
state\_action\_critic\_2.fc3.weight                 & {[}128,128{]}  \\
state\_action\_critic\_2.fc3.bias                   & {[}128{]}      \\
state\_action\_critic\_2.q.weight                   & {[}1,128{]}    \\
state\_action\_critic\_2.q.bias                     & {[}1{]}        \\
target\_state\_critic.fc1.weight                    & {[}128,128{]}  \\
target\_state\_critic.fc1.bias                      & {[}128{]}      \\
target\_state\_critic.fc2.weight                    & {[}128,128{]}  \\
target\_state\_critic.fc2.bias                      & {[}128{]}      \\
target\_state\_critic.fc3.weight                    & {[}128,128{]}  \\
target\_state\_critic.fc3.bias                      & {[}128{]}      \\
target\_state\_critic.v.weight                      & {[}1,128{]}    \\
target\_state\_critic.v.bias                        & {[}1{]}        \\
template\_decoder\_network.tmpl\_gru.weight\_ih\_l0 & {[}384,128{]}  \\
template\_decoder\_network.tmpl\_gru.weight\_hh\_l0 & {[}384,128{]}  \\
template\_decoder\_network.tmpl\_gru.bias\_ih\_l0   & {[}384{]}      \\
template\_decoder\_network.tmpl\_gru.bias\_hh\_l0   & {[}384{]}      \\
template\_decoder\_network.fc2.weight               & {[}128,128{]}  \\
template\_decoder\_network.fc2.bias                 & {[}128{]}      \\
template\_decoder\_network.tmpl.weight              & {[}235,128{]}  \\
template\_decoder\_network.tmpl.bias                & {[}235{]}      \\
object\_decoder\_network.obj\_gru.weight\_ih\_l0    & {[}384,256{]}  \\
object\_decoder\_network.obj\_gru.weight\_hh\_l0    & {[}384,128{]}  \\
object\_decoder\_network.obj\_gru.bias\_ih\_l0      & {[}384{]}      \\
object\_decoder\_network.obj\_gru.bias\_hh\_l0      & {[}384{]}      \\
object\_decoder\_network.fc2.weight                 & {[}128,128{]}  \\
object\_decoder\_network.fc2.bias                   & {[}128{]}      \\
object\_decoder\_network.obj.weight                 & {[}699,128{]}  \\
object\_decoder\_network.obj.bias                   & {[}699{]}      \\ \hline
\end{tabular}
}
\label{tab:parameter-size}
\end{table}

%% file: table/tab-trn-time.tex
\begin{table}[ht!]
\centering
\scriptsize
\begin{tabular}{c|ccc}
\hline
       & Step/second (CPU)    & Step/second (GPU)    & Parameter Size \\ \hline
KG-A2C & 0.28                  & 0.71                  & 4.8M           \\
TAC    & 0.99                  & 1.43                  & 1.8M           \\ \hline
\end{tabular}
\caption{Training time as step per second in CPU and GPU and total parameter size for \gzorko.}
\label{tab:training-time-hyper-parameters}
\end{table}

%% file: table/tab-qual1-case.tex
\begin{table*}[t]
\centering
\tiny
\begin{tabularx}{16cm}{X}
\hline
\multicolumn{1}{c}{\texttt{Case 1.1}}                                                                                                                                                                                                                                                                                                                                                                                                                                                                                                                                                                                                                                                                                                                                                                                                                                                                                                                                                                                                                                            \\ \hline
\texttt{Step: 4 \newline
Game: Kitchen You are in the kitchen of the white house. A table seems to have been used recently for the preparation of food. A passage leads to the west and a dark staircase can be seen leading upward. A dark chimney leads down and to the east is a small window which is open. On the table is an elongated brown sack, smelling of hot peppers. A bottle is sitting on the table. The glass bottle contains: A quantity of water \newline
Look: Kitchen You are in the kitchen of the white house. A table seems to have been used recently for the preparation of food. A passage leads to the west and a dark staircase can be seen leading upward. A dark chimney leads down and to the east is a small window which is open. On the table is an elongated brown sack, smelling of hot peppers. A bottle is sitting on the table. The glass bottle contains: A quantity of water \newline
Inv: You are empty handed. \newline
Score: 10 \newline
Action: west} \\ \hline
\multicolumn{1}{c}{\texttt{Case 1.2}}                                                                                                                                                                                                                                                                                                                                                                                                                                                                                                                                                                                                                                                                                                                                                                                                                                                                                                                                                                                                                                            \\ \hline
\texttt{Step: 15 \newline
Game: Kitchen On the table is an elongated brown sack, smelling of hot peppers. A bottle is sitting on the table. The glass bottle contains: A quantity of water \newline
Look: Kitchen You are in the kitchen of the white house. A table seems to have been used recently for the preparation of food. A passage leads to the west and a dark staircase can be seen leading upward. A dark chimney leads down and to the east is a small window which is open. On the table is an elongated brown sack, smelling of hot peppers. A bottle is sitting on the table. The glass bottle contains: A quantity of water \newline
Inv: You are carrying: A painting A brass lantern (providing light) \newline
Score: 39 \newline
Action: west}                                                                                                                                                                                                                                \\ \hline
\multicolumn{1}{c}{\texttt{Case 1.3}}                                                                                                                                                                                                                                                                                                                                                                                                                                                                                                                                                                                                                                                                                                                                                                                                                                                                                                                                                                                                                                            \\ \hline
\texttt{Step: 20 \newline
Game: Kitchen On the table is an elongated brown sack, smelling of hot peppers. A bottle is sitting on the table. The glass bottle contains: A quantity of water \newline
Look: Kitchen You are in the kitchen of the white house. A table seems to have been used recently for the preparation of food. A passage leads to the west and a dark staircase can be seen leading upward. A dark chimney leads down and to the east is a small window which is open. On the table is an elongated brown sack, smelling of hot peppers. A bottle is sitting on the table. The glass bottle contains: A quantity of water \newline
Inv: You are empty handed. \newline
Score: 45 \newline
Action: east}                                                                                                                                                                                                                                                                        \\ \hline
\end{tabularx}
\caption{Case 1; Game observation and the selected action snippets from \gzorko.}
\label{tab:qual1-case}
\end{table*}

%% file: table/tab-qual1-anal.tex
\begin{table*}[t]
\centering
\scriptsize
\begin{tabular}{lcccccc}
\hline
\multicolumn{7}{c}{\texttt{Case 1.1}}                                                                                                                                                                                    \\ \hline
\multicolumn{1}{l|}{}     & \multicolumn{2}{c|}{$ n_{\text{score}} = 10 $}                       & \multicolumn{2}{c|}{$ n_{\text{score}} = 39 $}                      & \multicolumn{2}{c}{$ n_{\text{score}} = 45 $}   \\
\multicolumn{1}{l|}{}     & $ \pi (a_{\mathbb{T}} | o) $ & \multicolumn{1}{c|}{$ Q (o, a) $}     & $ \pi (a_{\mathbb{T}} | o) $ & \multicolumn{1}{c|}{$ Q (o, a) $}    & $ \pi (a_{\mathbb{T}} | o) $  & $ Q (o, a) $    \\ \hline
\multicolumn{1}{l|}{west} & \textbf{0.9998}              & \multicolumn{1}{c|}{\textbf{23.7460}} & 0.000                        & \multicolumn{1}{c|}{4.1434}          & 0.000                         & 5.0134          \\
\multicolumn{1}{l|}{east} & 0.000                        & \multicolumn{1}{c|}{18.4385}          & \textbf{0.5674}              & \multicolumn{1}{c|}{\textbf{5.1640}} & \textbf{0.9996}               & \textbf{6.0319} \\ \hline
\multicolumn{7}{c}{\texttt{Case 1.2}}                                                                                                                                                                                    \\ \hline
\multicolumn{1}{l|}{}     & \multicolumn{2}{c|}{$ n_{\text{score}} = 10 $}                       & \multicolumn{2}{c|}{$ n_{\text{score}} = 39 $}                      & \multicolumn{2}{c}{$ n_{\text{score}} = 45 $}   \\
\multicolumn{1}{l|}{}     & $ \pi (a_{\mathbb{T}} | o) $ & \multicolumn{1}{c|}{$ Q (o, a) $}     & $ \pi (a_{\mathbb{T}} | o) $ & \multicolumn{1}{c|}{$ Q (o, a) $}    & $ \pi (a_{\mathbb{T}} | o) $  & $ Q (o, a) $    \\ \hline
\multicolumn{1}{l|}{west} & \textbf{0.9975}              & \multicolumn{1}{c|}{\textbf{27.6005}} & \textbf{0.9819}              & \multicolumn{1}{c|}{\textbf{8.3794}} & \textbf{0.8967}               & \textbf{8.0586} \\
\multicolumn{1}{l|}{east} & 0.000                        & \multicolumn{1}{c|}{23.6015}          & 0.0002                       & \multicolumn{1}{c|}{6.5284}          & 0.000                         & 6.4848          \\ \hline
\multicolumn{7}{c}{\texttt{Case 1.3}}                                                                                                                                                                                    \\ \hline
\multicolumn{1}{l|}{}     & \multicolumn{2}{c|}{$ n_{\text{score}} = 10 $}                       & \multicolumn{2}{c|}{$ n_{\text{score}} = 39 $}                      & \multicolumn{2}{c}{$ n_{\text{score}} = 45 $}   \\
\multicolumn{1}{l|}{}     & $ \pi (a_{\mathbb{T}} | o) $ & \multicolumn{1}{c|}{$ Q (o, a) $}     & $ \pi (a_{\mathbb{T}} | o) $ & \multicolumn{1}{c|}{$ Q (o, a) $}    & $ \pi (a_{\mathbb{T}} | o) $  & $ Q (o, a) $    \\ \hline
\multicolumn{1}{l|}{west} & \textbf{0.7872}              & \multicolumn{1}{c|}{\textbf{22.2419}} & 0.0001                       & \multicolumn{1}{c|}{4.9664}          & 0.000                         & 5.0169          \\
\multicolumn{1}{l|}{east} & 0.0055                       & \multicolumn{1}{c|}{19.1751}          & \textbf{0.7821}              & \multicolumn{1}{c|}{\textbf{5.7299}} & \textbf{0.9999}               & \textbf{6.2653} \\ \hline
\end{tabular}
\caption{Case 1; The changes in policy and Q value based on the score embedding from \gzorko.}
\label{tab:qual1-anal}
\end{table*}

%% file: table/tab-qual2-case.tex
\begin{table*}[t]
\centering
\tiny
\begin{tabularx}{16cm}{X}
\hline
\multicolumn{1}{c}{\texttt{Case 2.1}}                                                                                                                                                                                                                                                                                                                                                                                                                                                                                                           \\ \hline
\texttt{Step: 2 \newline
Game: Behind House You are behind the white house. A path leads into the forest to the east. In one corner of the house there is a small window which is slightly ajar. \newline
Look: Behind House You are behind the white house. A path leads into the forest to the east. In one corner of the house there is a small window which is slightly ajar. \newline
Inv: You are empty handed. \newline
Score: 0 \newline
Action: open window} \\ \hline
\multicolumn{1}{c}{\texttt{Case 2.2}}                                                                                                                                                                                                                                                                                                                                                                                                                                                                                                           \\ \hline
\texttt{Step: 3 \newline
Game: With great effort, you open the window far enough to allow entry. \newline
Look: Behind House You are behind the white house. A path leads into the forest to the east. In one corner of the house there is a small window which is open. \newline
Inv: You are empty handed. \newline
Score: 0 \newline
Action: west}                                                                                                                \\ \hline
\multicolumn{1}{c}{\texttt{Case 2.3}}                                                                                                                                                                                                                                                                                                                                                                                                                                                                                                           \\ \hline
\texttt{Step: 21 \newline
Game: Behind House \newline
Look: Behind House You are behind the white house. A path leads into the forest to the east. In one corner of the house there is a small window which is open. \newline
Inv: You are empty handed. \newline
Score: 45 \newline
Action: north}                                                                                                                                                                  \\ \hline
\end{tabularx}
\caption{Case 2; Game observation and the selected action snippets from \gzorko.}
\label{tab:qual2-case}
\end{table*}

%% file: table/tab-qual2-anal.tex
\begin{table*}[t]
\centering
\scriptsize
\begin{tabular}{lcccc}
\hline
\multicolumn{5}{c}{\texttt{Case 2.1}}                                                                                                                     \\ \hline
\multicolumn{1}{l|}{}            & \multicolumn{2}{c|}{$ n_{\text{score}} = 0 $}                        & \multicolumn{2}{c}{$ n_{\text{score}} = 45 $}   \\
\multicolumn{1}{l|}{}            & $ \pi (a_{\mathbb{T}} | o) $ & \multicolumn{1}{c|}{$ Q (o, a) $}     & $ \pi (a_{\mathbb{T}} | o) $  & $ Q (o, a) $    \\ \hline
\multicolumn{1}{l|}{open window} & \textbf{0.9999}              & \multicolumn{1}{c|}{\textbf{29.0205}} & 0.0111                        & 5.9599          \\
\multicolumn{1}{l|}{west}        & 0.0000                       & \multicolumn{1}{c|}{28.6848}          & 0.0893                        & 6.1119          \\
\multicolumn{1}{l|}{north}       & 0.0000                       & \multicolumn{1}{c|}{26.7997}          & \textbf{0.8174}               & \textbf{6.2819} \\ \hline
\multicolumn{5}{c}{\texttt{Case 2.2}}                                                                                                                     \\ \hline
\multicolumn{1}{l|}{}            & \multicolumn{2}{c|}{$ n_{\text{score}} = 0 $}                        & \multicolumn{2}{c}{$ n_{\text{score}} = 45 $}   \\
\multicolumn{1}{l|}{}            & $ \pi (a_{\mathbb{T}} | o) $ & \multicolumn{1}{c|}{$ Q (o, a) $}     & $ \pi (a_{\mathbb{T}} | o) $  & $ Q (o, a) $    \\ \hline
\multicolumn{1}{l|}{open window} & 0.0000                       & \multicolumn{1}{c|}{30.2154}          & 0.0000                        & 6.1354          \\
\multicolumn{1}{l|}{west}        & \textbf{0.9999}              & \multicolumn{1}{c|}{\textbf{32.0298}} & 0.0000                        & 5.8312          \\
\multicolumn{1}{l|}{north}       & 0.0000                       & \multicolumn{1}{c|}{26.7509}          & \textbf{0.9952}               & \textbf{6.6669} \\ \hline
\multicolumn{5}{c}{\texttt{Case 2.3}}                                                                                                                     \\ \hline
\multicolumn{1}{l|}{}            & \multicolumn{2}{c|}{$ n_{\text{score}} = 0 $}                        & \multicolumn{2}{c}{$ n_{\text{score}} = 45 $}   \\
\multicolumn{1}{l|}{}            & $ \pi (a_{\mathbb{T}} | o) $ & \multicolumn{1}{c|}{$ Q (o, a) $}     & $ \pi (a_{\mathbb{T}} | o) $  & $ Q (o, a) $    \\ \hline
\multicolumn{1}{l|}{open window} & 0.0000                       & \multicolumn{1}{c|}{30.2184}          & 0.0001                        & 6.0443          \\
\multicolumn{1}{l|}{west}        & \textbf{0.9999}              & \multicolumn{1}{c|}{\textbf{32.0302}} & 0.0000                        & 5.6724          \\
\multicolumn{1}{l|}{north}       & 0.0000                       & \multicolumn{1}{c|}{26.7494}          & \textbf{0.9867}               & \textbf{6.5545} \\ \hline
\end{tabular}
\caption{Case 2; The changes in policy and Q value based on the score embedding from \gzorko.}
\label{tab:qual2-anal}
\end{table*}

%% file: figure/tac-main.tex
\begin{figure*}[htbp!]
    \centering
    \includegraphics[width=0.8\textwidth]{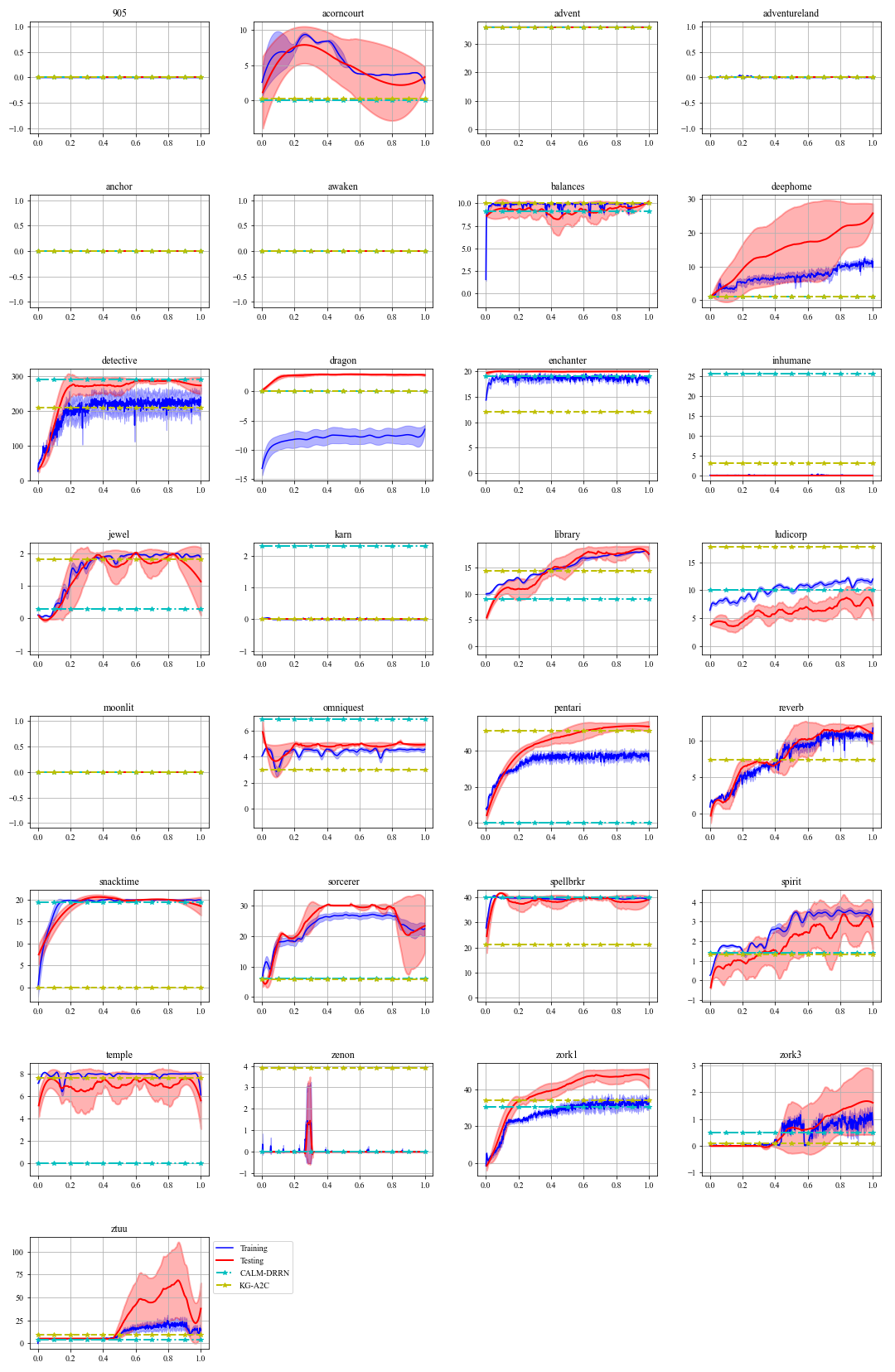}
    \caption{The full learning curve for TAC, compared with TDQN and KG-A2C}
    \label{fig:tac-main}
\end{figure*}

%% file: table/tab-main-all.tex
\begin{table*}[ht!]
\centering
\footnotesize
\begin{tabular}{l|ccc|ccc}
\hline
                        & NAIL          & DRRN          & TDQN          & CALM-DRRN               & KG-A2C                 & TAC                         \\ \hline
\gnof                   & 0.0           & 0.0           & 0.0           & 0.0                     & 0.0                    & 0.0 $ \pm $ 0.0             \\
\gacorncourt            & 0.0           & \textit{10.0} & 1.6           & 0.0                     & 0.3                    & \textbf{3.4 $ \pm $ 1.6}    \\
\gadvent$^{\dagger}$    & 36.0          & 36.0          & 36.0          & 36.0                    & 36.0                   & 36.0 $ \pm $ 0.0            \\
\gadventureland         & 0.0           & \textit{20.6} & 0.0           & 0.0                     & 0.0                    & 0.0 $ \pm $ 0.0             \\
\ganchor                & 0.0           & 0.0           & 0.0           & 0.0                     & 0.0                    & 0.0 $ \pm $ 0.0             \\
\gawaken                & 0.0           & 0.0           & 0.0           & 0.0                     & 0.0                    & 0.0 $ \pm $ 0.0             \\
\gbalances              & \textit{10.0} & \textit{10.0} & 4.8           & 9.1                     & \textit{\textbf{10.0}} & \textit{\textbf{10.0 $ \pm $ 0.1}}   \\
\gdeephome$^{\ddagger}$ & 13.3          & 1.0           & 1.0           & 1.0                     & 1.0                    & \textit{\textbf{25.4 $ \pm $ 3.2}}   \\
\gdetective             & 136.9         & 197.8         & 169.0         & \textit{\textbf{289.7}} & 207.9                  & 272.3 $ \pm $ 23.3          \\
\gdragon                & 0.6           & -3.5          & -5.3          & 0.1                     & 0.0                    & \textit{\textbf{2.81 $ \pm $ 0.15}}  \\
\genchanter             & 0.0           & \textit{20.0} & 8.6           & 19.1                    & 12.1                   & \textit{\textbf{20.0 $ \pm $ 0.0}}   \\
\ginhumane              & 0.6           & 0.0           & 0.7           & \textit{\textbf{25.7}}  & 3.0                    & 0.0 $ \pm $ 0.0             \\
\gjewel                 & 1.6           & 1.6           & 0.0           & 0.3                     & \textit{\textbf{1.8}}  & 1.17 $ \pm $ 1.0            \\
\gkarn                  & 1.2           & 2.1           & 0.7           & \textit{\textbf{2.3}}   & 0.0                    & 0.0 $ \pm $ 0.0             \\
\glibrary               & 0.9           & 17.0          & 6.3           & 9.0                     & 14.3                   & \textit{\textbf{18.0 $ \pm $ 1.2}}   \\
\gludicorp              & 8.4           & 13.8          & 6.0           & 10.1                    & \textit{\textbf{17.8}} & 7.7 $ \pm $ 2.5             \\
\gmoonlit               & 0.0           & 0.0           & 0.0           & 0.0                     & 0.0                    & 0.0 $ \pm $ 0.0             \\
\gomniquest             & 5.6           & 5.0           & \textit{16.8} & \textbf{6.9}            & 3.0                    & 4.9 $ \pm $ 0.1             \\
\gpentari               & 0.0           & 27.2          & 17.4          & 0.0                     & 50.7                   & \textit{\textbf{53.2 $ \pm $ 2.9}}   \\
\greverb                & 0.0           & 8.2           & 0.3           & $-$                     & 7.4                    & \textit{\textbf{11 $ \pm $ 1.4}}     \\
\gsnacktime             & 0.0           & 0.0           & 9.7           & \textit{\textbf{19.4}}  & 0.0                    & 18.6 $ \pm $ 2.0   \\
\gsorcerer              & 5.0           & 20.8          & 5.0           & 6.2                     & 5.8                    & \textit{\textbf{23.2 $ \pm $ 9.3}}   \\
\gspellbrkr             & \textit{40.0} & 37.8          & 18.7          & \textit{\textbf{40.0}}  & 21.3                   & 39.0 $ \pm $ 1.4            \\
\gspirit                & 1.0           & 0.8           & 0.6           & 1.4                     & 1.3                    & \textit{\textbf{2.91 $ \pm $ 1.1}}   \\
\gtemple                & 7.3           & 7.4           & \textit{7.9}  & 0.0                     & \textbf{7.6}           & 5.8 $ \pm $ 2.3             \\
\gzenon                 & 0.0           & 0.0           & 0.0           & 0.0                     & \textit{\textbf{3.9}}  & 0.0 $ \pm $ 0.0             \\
\gzorko                 & 10.3          & 32.6          & 9.9           & 30.4                    & 34                     & \textit{\textbf{46.3 $ \pm $ 5.0}}   \\
\gzorkt                 & \textit{1.8}  & 0.5           & 0.0           & 0.5                     & 0.1                    & \textbf{1.6 $ \pm $ 1.2}             \\
\gztuu                  & 0.0           & 21.6          & 4.9           & 3.7                     & 9.2                    & \textit{\textbf{33.2 $ \pm $ 26.0}}  \\ \hline
MEAN                    & 0.0536        & 0.1156        & 0.0665        & 0.0936                  & 0.1094                 & \textit{\textbf{0.1560}}             \\ \hline
\end{tabular}
\caption{Game score comparison over 29 game environments in Jericho, with best results highlighted by \textbf{boldface}. NAIL and DRRN are non-generative baselines while TDQN and KG-A2C are generative baselines. The last row is the mean game score over all the environments. The initial game score of \gadvent$^\dagger$ is 36 and \gdeephome$^{\ddagger}$ is 1.}
\label{tab:main-all}
\end{table*}

%% file: figure/tac-regularization-zork1.tex
\begin{figure}[t]
    \centering
    \includegraphics[width=0.4\textwidth]{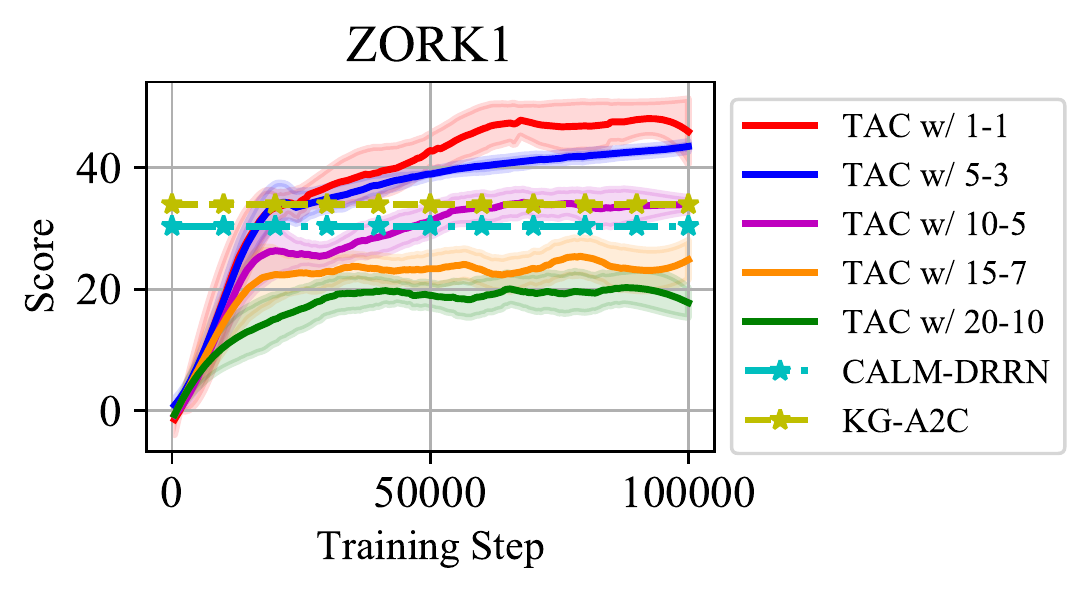}
    \caption{The learning curve of TAC for \textit{regularization} ablation in \gzorko. Stronger supervised signals are used with $ \epsilon = 0.3 $, where 5-3 signifies $ \gamma_{ \mathbb{T} } = 5 $ and $ \gamma_{ \mathbb{O} } = 3 $.}
    \label{fig:tac-regularization-zork1}
\end{figure}


%% file: figure/tac-dynamic-eps-1.tex
\begin{figure*}[t!]
    \centering
    \includegraphics[width=1\textwidth]{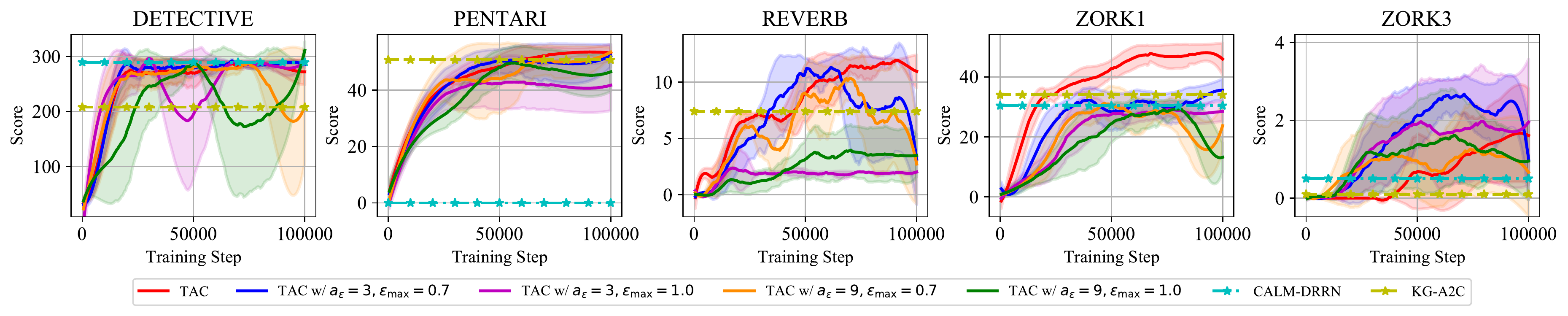}
    \caption{The learning curve of TAC with dynamic epsilon on five popular games. All the experiments were done with fixed $ \epsilon_{\text{min}} = 0.0 $, $ a_{\epsilon} = \{ 3 , 9 \} $ and $ \epsilon_{\text{max}} = \{ 0.7 , 1.0 \} $.}
    \label{fig:tac-dynamic-eps-1}
\end{figure*}

%% file: figure/tac-dynamic-eps-2.tex
\begin{figure*}[t!]
    \centering
    \includegraphics[width=1\textwidth]{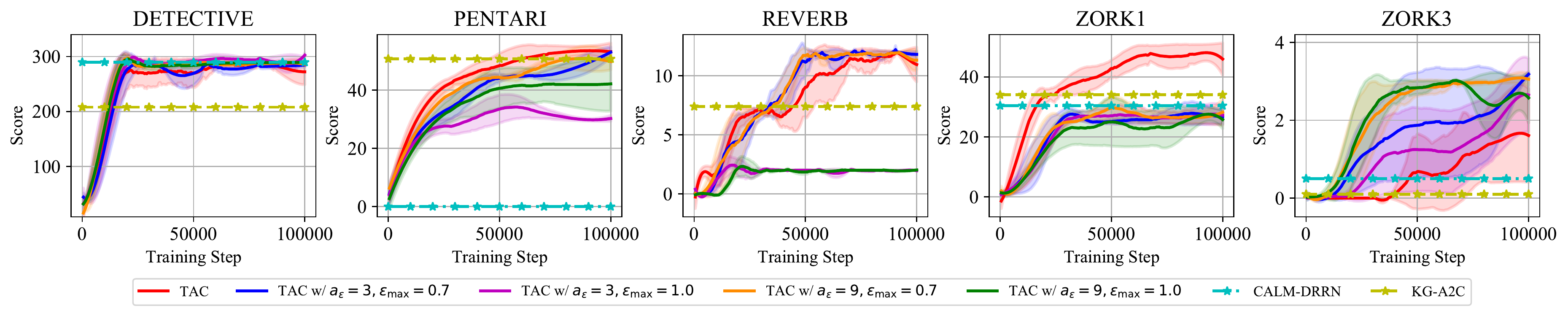}
    \caption{The learning curve of TAC with dynamic epsilon on five popular games. All the experiments were done with fixed $ \epsilon_{\text{min}} = 0.3 $, $ a_{\epsilon} = \{ 3 , 9 \} $ and $ \epsilon_{\text{max}} = \{ 0.7 , 1.0 \} $.}
    \label{fig:tac-dynamic-eps-2}
\end{figure*}

%% file: figure/scr2prob.tex
\begin{figure}[t!]
    \centering
    \includegraphics[width=0.45\textwidth]{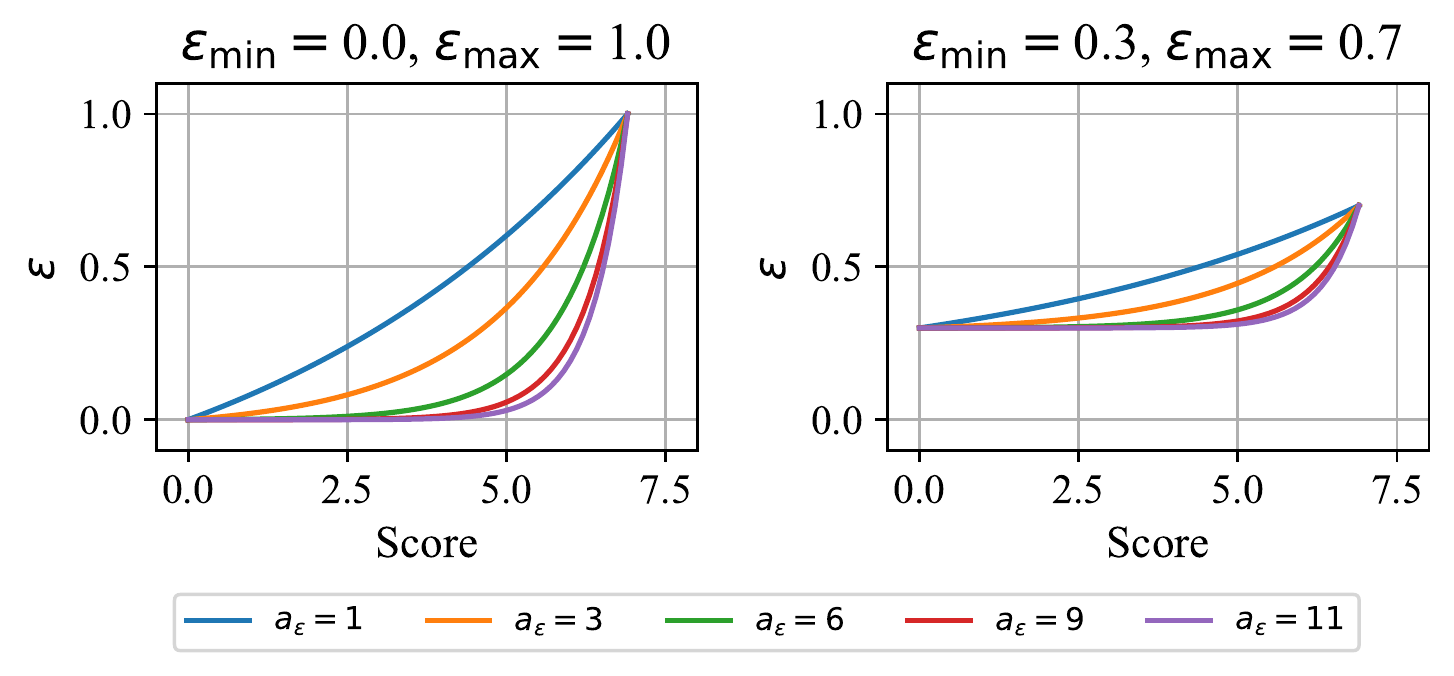}
    \caption{The exponential probability of $ \epsilon $ over the game score. Left is with $ \epsilon_{\min} = 0.0,\epsilon_{\max} = 1.0 $ and right is with $ \epsilon_{\min} = 0.3,\epsilon_{\max} = 0.7 $ between the game score of 0 to 6. Five different $ a_{\epsilon} $ is drawn per plot.}
    \label{fig:scr2prob}
\end{figure}

%% file: data/zork1.tex
\begin{lstlisting}[breaklines,basicstyle=\tiny]
=== STEP 0 ===
Game: Copyright (c) 1981, 1982, 1983 Infocom, Inc. All rights reserved. ZORK is a registered trademark of Infocom, Inc. Revision 88 / Serial number 840726  West of House You are standing in an open field west of a white house, with a boarded front door. There is a small mailbox here.
Look: West of House You are standing in an open field west of a white house, with a boarded front door. There is a small mailbox here.
Inv: You are empty handed.
State Value: 25.94228
Top 20 Templates: north (0.64085) | south (0.35826) | go around OBJ (0.00061) | northwest (0.0001) | put OBJ in OBJ (8e-05) | put down OBJ (3e-05) | exit (2e-05) | turn on OBJ (1e-05) | southwest (1e-05) | close OBJ (0.0) | pull OBJ (0.0) | hit OBJ with OBJ (0.0) | shake OBJ (0.0) | open OBJ (0.0) | open OBJ with OBJ (0.0) | down (0.0) | west (0.0) | give OBJ to OBJ (0.0) | wait (0.0) | up (0.0)
Q Value for Valid Action: open mailbox (24.6124, 24.71131) | north (26.11693, 26.19618) | south (26.125, 26.17242) | west (22.82211, 22.94841)
Action: (south, air, hide) => south (Q Value: 26.125)
Reward: 0
Score: 0
If Valid: True

=== STEP 1 ===
Game: South of House You are facing the south side of a white house. There is no door here, and all the windows are boarded.
Look: South of House You are facing the south side of a white house. There is no door here, and all the windows are boarded.
Inv: You are empty handed.
State Value: 27.86528
Top 20 Templates: east (0.99991) | go around OBJ (6e-05) | up (1e-05) | exit (1e-05) | south (0.0) | pull OBJ (0.0) | put down OBJ (0.0) | west (0.0) | north (0.0) | northwest (0.0) | put OBJ in OBJ (0.0) | blow out OBJ (0.0) | shake OBJ (0.0) | close OBJ (0.0) | jump (0.0) | turn on OBJ (0.0) | southwest (0.0) | give OBJ to OBJ (0.0) | open OBJ with OBJ (0.0) | put OBJ on OBJ (0.0)
Q Value for Valid Action: west (24.71801, 24.7743) | south (24.22558, 24.54047) | east (27.94924, 28.01769)
Action: (east, knife, bloody) => east (Q Value: 27.94924)
Reward: 0
Score: 0
If Valid: True

=== STEP 2 ===
Game: Behind House You are behind the white house. A path leads into the forest to the east. In one corner of the house there is a small window which is slightly ajar.
Look: Behind House You are behind the white house. A path leads into the forest to the east. In one corner of the house there is a small window which is slightly ajar.
Inv: You are empty handed.
State Value: 29.34139
Top 20 Templates: open OBJ (0.99969) | turn on OBJ (0.00017) | take all (0.0001) | close OBJ (1e-05) | shake OBJ (1e-05) | put down OBJ (0.0) | put out OBJ (0.0) | north (0.0) | hit OBJ with OBJ (0.0) | take OBJ (0.0) | fill OBJ with OBJ (0.0) | put OBJ in OBJ (0.0) | south (0.0) | push OBJ (0.0) | look at OBJ with OBJ (0.0) | pour OBJ from OBJ (0.0) | east (0.0) | turn on OBJ with OBJ (0.0) | exit (0.0) | open OBJ with OBJ (0.0)
Q Value for Valid Action: open window (29.57216, 29.55614) | south (26.45868, 26.59624) | north (26.46529, 26.50062) | east (26.78434, 26.89982)
Action: (open OBJ, small, egg) => open small (Q Value: 29.56306)
Reward: 0
Score: 0
If Valid: True

=== STEP 3 ===
Game: With great effort, you open the window far enough to allow entry.
Look: Behind House You are behind the white house. A path leads into the forest to the east. In one corner of the house there is a small window which is open.
Inv: You are empty handed.
State Value: 31.49152
Top 20 Templates: west (0.99987) | take all (6e-05) | open OBJ (3e-05) | take OBJ (2e-05) | turn on OBJ (0.0) | northwest (0.0) | north (0.0) | close OBJ (0.0) | south (0.0) | put out OBJ (0.0) | east (0.0) | blow out OBJ (0.0) | throw OBJ at OBJ (0.0) | shake OBJ (0.0) | go around OBJ (0.0) | put OBJ on OBJ (0.0) | wait (0.0) | jump (0.0) | hit OBJ with OBJ (0.0) | turn on OBJ with OBJ (0.0)
Q Value for Valid Action: west (31.65153, 31.64213) | close window (27.66837, 27.62293) | south (26.53292, 26.4846) | north (26.61803, 26.44608) | east (26.87151, 26.82128)
Action: (west, egg, egg) => west (Q Value: 31.64213)
Reward: 10
Score: 10
If Valid: True

=== STEP 4 ===
Game: Kitchen You are in the kitchen of the white house. A table seems to have been used recently for the preparation of food. A passage leads to the west and a dark staircase can be seen leading upward. A dark chimney leads down and to the east is a small window which is open. On the table is an elongated brown sack, smelling of hot peppers. A bottle is sitting on the table. The glass bottle contains:   A quantity of water
Look: Kitchen You are in the kitchen of the white house. A table seems to have been used recently for the preparation of food. A passage leads to the west and a dark staircase can be seen leading upward. A dark chimney leads down and to the east is a small window which is open. On the table is an elongated brown sack, smelling of hot peppers. A bottle is sitting on the table. The glass bottle contains:   A quantity of water
Inv: You are empty handed.
State Value: 23.26802
Top 20 Templates: west (0.99984) | northwest (3e-05) | close OBJ (3e-05) | push OBJ (2e-05) | south (1e-05) | blow out OBJ (1e-05) | exit (1e-05) | put OBJ on OBJ (1e-05) | go around OBJ (0.0) | put OBJ in OBJ (0.0) | open OBJ (0.0) | turn on OBJ (0.0) | take all (0.0) | north (0.0) | take OBJ (0.0) | shake OBJ (0.0) | east (0.0) | up (0.0) | jump (0.0) | put out OBJ (0.0)
Q Value for Valid Action: jump (-10.0567, -9.87989) | east (17.97866, 18.21377) | open sack (21.87809, 21.87084) | open bottle (21.61428, 21.67213) | take sack (20.64789, 20.65431) | take bottle (20.71271, 20.84131) | up (18.40112, 18.3842) | close window (21.18477, 21.23618) | west (23.79584, 23.77861) | take all (19.97321, 20.06381)
Action: (west, glue, glue) => west (Q Value: 23.77861)
Reward: 0
Score: 10
If Valid: True

=== STEP 5 ===
Game: Living Room You are in the living room. There is a doorway to the east, a wooden door with strange gothic lettering to the west, which appears to be nailed shut, a trophy case, and a large oriental rug in the center of the room. Above the trophy case hangs an elvish sword of great antiquity. A battery powered brass lantern is on the trophy case.
Look: Living Room You are in the living room. There is a doorway to the east, a wooden door with strange gothic lettering to the west, which appears to be nailed shut, a trophy case, and a large oriental rug in the center of the room. Above the trophy case hangs an elvish sword of great antiquity. A battery powered brass lantern is on the trophy case.
Inv: You are empty handed.
State Value: 25.1383
Top 20 Templates: push OBJ (0.99748) | take all (0.00212) | fill OBJ with OBJ (0.0001) | west (8e-05) | put OBJ on OBJ (6e-05) | northwest (5e-05) | exit (2e-05) | turn on OBJ (2e-05) | open OBJ (1e-05) | put down OBJ (1e-05) | shake OBJ (1e-05) | close OBJ (1e-05) | take OBJ from OBJ (1e-05) | blow out OBJ (1e-05) | take OBJ (0.0) | down (0.0) | pour OBJ from OBJ (0.0) | look at OBJ with OBJ (0.0) | put out OBJ (0.0) | throw OBJ at OBJ (0.0)
Q Value for Valid Action: turn on brass (23.94045, 23.92543) | open case (23.69346, 23.7169) | take brass (25.17443, 25.15079) | take elvish (23.20454, 23.19269) | push oriental (25.46935, 25.5534) | east (19.64956, 19.7294) | take all (24.79437, 24.78484)
Action: (push OBJ, orient, glue) => push orient (Q Value: 25.46935)
Reward: 0
Score: 10
If Valid: True

=== STEP 6 ===
Game: With a great effort, the rug is moved to one side of the room, revealing the dusty cover of a closed trap door.
Look: Living Room You are in the living room. There is a doorway to the east, a wooden door with strange gothic lettering to the west, which appears to be nailed shut, a trophy case, and a closed trap door at your feet. Above the trophy case hangs an elvish sword of great antiquity. A battery powered brass lantern is on the trophy case.
Inv: You are empty handed.
State Value: 27.01657
Top 20 Templates: open OBJ (0.99907) | take all (0.00077) | take OBJ from OBJ (3e-05) | fill OBJ with OBJ (3e-05) | up (2e-05) | put out OBJ (1e-05) | turn on OBJ (1e-05) | put OBJ on OBJ (1e-05) | put down OBJ (1e-05) | take OBJ (1e-05) | close OBJ (1e-05) | open OBJ with OBJ (0.0) | blow out OBJ (0.0) | push OBJ (0.0) | shake OBJ (0.0) | look at OBJ with OBJ (0.0) | exit (0.0) | throw OBJ at OBJ (0.0) | pour OBJ from OBJ (0.0) | put OBJ in OBJ (0.0)
Q Value for Valid Action: turn on brass (25.7593, 25.63789) | open case (25.34084, 25.45546) | open cover (27.26838, 27.29786) | take antiquity (25.06986, 25.06573) | take brass (27.0097, 27.02397) | east (19.36988, 19.32297) | take all (26.59528, 26.80509)
Action: (open OBJ, trap, sand) => open trap (Q Value: 27.26035)
Reward: 0
Score: 10
If Valid: True

=== STEP 7 ===
Game: The door reluctantly opens to reveal a rickety staircase descending into darkness.
Look: Living Room You are in the living room. There is a doorway to the east, a wooden door with strange gothic lettering to the west, which appears to be nailed shut, a trophy case, and a rug lying beside an open trap door. Above the trophy case hangs an elvish sword of great antiquity. A battery powered brass lantern is on the trophy case.
Inv: You are empty handed.
State Value: 29.02823
Top 20 Templates: take OBJ (0.89478) | take all (0.10472) | open OBJ with OBJ (0.00022) | shake OBJ (6e-05) | put down OBJ (5e-05) | open OBJ (3e-05) | turn on OBJ (3e-05) | put out OBJ (2e-05) | take OBJ from OBJ (2e-05) | blow out OBJ (1e-05) | put OBJ on OBJ (1e-05) | fill OBJ with OBJ (1e-05) | push OBJ (1e-05) | take on OBJ (1e-05) | put OBJ in OBJ (0.0) | close OBJ (0.0) | pull OBJ (0.0) | wait (0.0) | up (0.0) | southwest (0.0)
Q Value for Valid Action: down (17.82736, 17.83106) | turn on brass (27.60222, 27.65736) | open case (27.34901, 27.28223) | take brass (29.3726, 29.3444) | take elvish (26.68949, 26.66518) | close trap (25.35298, 25.25046) | east (19.23231, 19.22583) | take all (28.8406, 28.86462)
Action: (take OBJ, brass, egg) => take brass (Q Value: 29.3444)
Reward: 0
Score: 10
If Valid: True

=== STEP 8 ===
Game: Taken.
Look: Living Room You are in the living room. There is a doorway to the east, a wooden door with strange gothic lettering to the west, which appears to be nailed shut, a trophy case, and a rug lying beside an open trap door. Above the trophy case hangs an elvish sword of great antiquity.
Inv: You are carrying:   A brass lantern
State Value: 30.8633
Top 20 Templates: down (0.99916) | turn on OBJ (0.00059) | fill OBJ with OBJ (6e-05) | take OBJ from OBJ (4e-05) | open OBJ with OBJ (3e-05) | shake OBJ (2e-05) | take on OBJ (2e-05) | open OBJ (2e-05) | put out OBJ (1e-05) | take OBJ (1e-05) | southwest (1e-05) | put down OBJ (0.0) | take all (0.0) | blow out OBJ (0.0) | close OBJ (0.0) | east (0.0) | give OBJ to OBJ (0.0) | throw OBJ at OBJ (0.0) | put OBJ in OBJ (0.0) | look around OBJ (0.0)
Q Value for Valid Action: down (30.64444, 30.62538) | turn on brass (29.95051, 29.95739) | open case (28.95129, 28.97396) | take elvish (28.51755, 28.36493) | close trap (26.6757, 26.74335) | put down brass (27.10902, 27.0532) | throw brass at case (17.46409, 17.3217) | east (25.65741, 25.56386)
Action: (down, sand, sand) => down (Q Value: 30.62538)
Reward: 25
Score: 35
If Valid: True

=== STEP 9 ===
Game: You have moved into a dark place. The trap door crashes shut, and you hear someone barring it.  It is pitch black. You are likely to be eaten by a grue.
Look: It is pitch black. You are likely to be eaten by a grue.
Inv: You are carrying:   A brass lantern
State Value: 6.17251
Top 20 Templates: turn on OBJ (0.97579) | close OBJ (0.01949) | open OBJ (0.00216) | take all (0.00084) | shake OBJ (0.00059) | put out OBJ (0.00025) | hit OBJ with OBJ (0.00022) | take OBJ (0.00016) | turn on OBJ with OBJ (0.00014) | pour OBJ from OBJ (9e-05) | look around OBJ (8e-05) | put OBJ in OBJ (4e-05) | southwest (4e-05) | look at OBJ with OBJ (3e-05) | give OBJ to OBJ (2e-05) | wait (2e-05) | fill OBJ with OBJ (1e-05) | give OBJ OBJ (1e-05) | take on OBJ (1e-05) | put OBJ on OBJ (0.0)
Q Value for Valid Action: east (-8.6414, -8.35889) | turn on lantern (6.9394, 7.08249) | put down lantern (-6.56797, -6.84492) | throw lantern at grue (-6.68143, -6.86579) | north (-9.66777, -9.59779) | south (-9.60609, -9.21734)
Action: (close OBJ, knife, knife) => close knife (Q Value: 5.92916)
Reward: 0
Score: 35
If Valid: False

=== STEP 10 ===
Game: It's too dark to see!
Look: It is pitch black. You are likely to be eaten by a grue.
Inv: You are carrying:   A brass lantern
State Value: 6.15705
Top 20 Templates: turn on OBJ (0.97547) | close OBJ (0.01987) | open OBJ (0.00206) | take all (0.00085) | shake OBJ (0.0006) | put out OBJ (0.00026) | hit OBJ with OBJ (0.00022) | take OBJ (0.00017) | turn on OBJ with OBJ (0.00014) | pour OBJ from OBJ (9e-05) | look around OBJ (8e-05) | put OBJ in OBJ (4e-05) | southwest (4e-05) | look at OBJ with OBJ (3e-05) | give OBJ to OBJ (2e-05) | wait (2e-05) | give OBJ OBJ (1e-05) | fill OBJ with OBJ (1e-05) | take on OBJ (1e-05) | put OBJ on OBJ (0.0)
Q Value for Valid Action: east (-8.70692, -8.09324) | turn on lantern (7.0266, 7.10819) | put down lantern (-6.60081, -6.67028) | throw lantern at grue (-6.59506, -6.70692) | north (-9.56126, -9.52402) | south (-9.60294, -9.27907)
Action: (close OBJ, knife, knife) => close knife (Q Value: 6.0499)
Reward: 0
Score: 35
If Valid: False

=== STEP 11 ===
Game: It's too dark to see!
Look: It is pitch black. You are likely to be eaten by a grue.
Inv: You are carrying:   A brass lantern
State Value: 6.15705
Top 20 Templates: turn on OBJ (0.97547) | close OBJ (0.01987) | open OBJ (0.00206) | take all (0.00085) | shake OBJ (0.0006) | put out OBJ (0.00026) | hit OBJ with OBJ (0.00022) | take OBJ (0.00017) | turn on OBJ with OBJ (0.00014) | pour OBJ from OBJ (9e-05) | look around OBJ (8e-05) | put OBJ in OBJ (4e-05) | southwest (4e-05) | look at OBJ with OBJ (3e-05) | give OBJ to OBJ (2e-05) | wait (2e-05) | give OBJ OBJ (1e-05) | fill OBJ with OBJ (1e-05) | take on OBJ (1e-05) | put OBJ on OBJ (0.0)
Q Value for Valid Action: east (-8.70692, -8.09324) | turn on lantern (7.0266, 7.10819) | put down lantern (-6.60081, -6.67028) | throw lantern at grue (-6.59506, -6.70692) | north (-9.56126, -9.52402) | south (-9.60294, -9.27907)
Action: (turn on OBJ, lanter, knife) => turn on lanter (Q Value: 7.0266)
Reward: 0
Score: 35
If Valid: True

=== STEP 12 ===
Game: The brass lantern is now on.  Cellar You are in a dark and damp cellar with a narrow passageway leading north, and a crawlway to the south. On the west is the bottom of a steep metal ramp which is unclimbable.
Look: Cellar You are in a dark and damp cellar with a narrow passageway leading north, and a crawlway to the south. On the west is the bottom of a steep metal ramp which is unclimbable.
Inv: You are carrying:   A brass lantern (providing light)
State Value: 7.46833
Top 20 Templates: south (0.99547) | east (0.00387) | go around OBJ (0.00033) | turn on OBJ (8e-05) | close OBJ (7e-05) | hit OBJ with OBJ (6e-05) | open OBJ (2e-05) | put OBJ in OBJ (2e-05) | put out OBJ (1e-05) | take OBJ (1e-05) | shake OBJ (1e-05) | pull OBJ (1e-05) | give OBJ to OBJ (1e-05) | southwest (1e-05) | put down OBJ (0.0) | look around OBJ (0.0) | up (0.0) | wait (0.0) | give OBJ OBJ (0.0) | throw OBJ at OBJ (0.0)
Q Value for Valid Action: put out lantern (6.39485, 6.25204) | put down lantern (6.22186, 6.21319) | throw lantern at dark (-5.67515, -5.70153) | north (6.39865, 6.17199) | south (8.03016, 8.02919)
Action: (south, bloody, hide) => south (Q Value: 8.02919)
Reward: 0
Score: 35
If Valid: True

=== STEP 13 ===
Game: East of Chasm You are on the east edge of a chasm, the bottom of which cannot be seen. A narrow passage goes north, and the path you are on continues to the east.
Look: East of Chasm You are on the east edge of a chasm, the bottom of which cannot be seen. A narrow passage goes north, and the path you are on continues to the east.
Inv: You are carrying:   A brass lantern (providing light)
State Value: 8.50237
Top 20 Templates: east (0.99935) | turn on OBJ (0.00021) | close OBJ (0.00015) | south (0.0001) | hit OBJ with OBJ (4e-05) | open OBJ (3e-05) | southwest (2e-05) | shake OBJ (2e-05) | go around OBJ (1e-05) | north (1e-05) | put out OBJ (1e-05) | look around OBJ (0.0) | put down OBJ (0.0) | west (0.0) | turn on OBJ with OBJ (0.0) | give OBJ to OBJ (0.0) | pull OBJ (0.0) | take OBJ (0.0) | look at OBJ with OBJ (0.0) | pour OBJ from OBJ (0.0)
Q Value for Valid Action: jump (-9.78649, -9.57489) | put out lantern (6.47491, 6.54967) | put down lantern (6.80327, 6.63812) | throw lantern at passage (-5.62236, -5.89553) | north (7.36888, 7.57795) | east (9.1861, 9.02587)
Action: (east, knife, knife) => east (Q Value: 9.02587)
Reward: 0
Score: 35
If Valid: True

=== STEP 14 ===
Game: Gallery This is an art gallery. Most of the paintings have been stolen by vandals with exceptional taste. The vandals left through either the north or west exits. Fortunately, there is still one chance for you to be a vandal, for on the far wall is a painting of unparalleled beauty.
Look: Gallery This is an art gallery. Most of the paintings have been stolen by vandals with exceptional taste. The vandals left through either the north or west exits. Fortunately, there is still one chance for you to be a vandal, for on the far wall is a painting of unparalleled beauty.
Inv: You are carrying:   A brass lantern (providing light)
State Value: 9.78262
Top 20 Templates: take OBJ (0.99217) | take all (0.00491) | up (0.00146) | open OBJ (0.00071) | turn on OBJ (0.00025) | shake OBJ (0.00015) | close OBJ (9e-05) | south (8e-05) | open OBJ with OBJ (7e-05) | put OBJ in OBJ (2e-05) | put out OBJ (2e-05) | southwest (1e-05) | wait (1e-05) | hit OBJ with OBJ (1e-05) | north (1e-05) | put down OBJ (1e-05) | west (1e-05) | go around OBJ (0.0) | look around OBJ (0.0) | pull OBJ (0.0)
Q Value for Valid Action: put out light (8.2289, 8.4294) | take paintings (10.24766, 10.31081) | put down light (8.16209, 8.13064) | throw light at paintings (0.57325, 0.84608) | north (7.78618, 7.93532) | west (8.12044, 8.10104)
Action: (take OBJ, painti, bloody) => take painti (Q Value: 10.24766)
Reward: 4
Score: 39
If Valid: True

=== STEP 15 ===
Game: Taken.
Look: Gallery This is an art gallery. Most of the paintings have been stolen by vandals with exceptional taste. The vandals left through either the north or west exits.
Inv: You are carrying:   A painting   A brass lantern (providing light)
State Value: 6.52417
Top 20 Templates: north (0.99797) | open OBJ (0.00116) | up (0.00042) | close OBJ (0.00012) | down (8e-05) | turn on OBJ (8e-05) | open OBJ with OBJ (5e-05) | southwest (3e-05) | put OBJ in OBJ (2e-05) | put out OBJ (1e-05) | shake OBJ (1e-05) | east (1e-05) | pull OBJ (1e-05) | northwest (1e-05) | take on OBJ (0.0) | wait (0.0) | blow out OBJ (0.0) | south (0.0) | put down OBJ (0.0) | go around OBJ (0.0)
Q Value for Valid Action: put out lantern (5.75779, 5.75869) | put down lantern (5.42923, 5.60296) | put down art (5.20534, 5.09223) | throw lantern at art (-2.01924, -2.64905) | north (7.37026, 7.20368) | west (5.56293, 5.3149)
Action: (north, glue, sand) => north (Q Value: 7.20368)
Reward: 0
Score: 39
If Valid: True

=== STEP 16 ===
Game: Studio This appears to have been an artist's studio. The walls and floors are splattered with paints of 69 different colors. Strangely enough, nothing of value is hanging here. At the south end of the room is an open door (also covered with paint). A dark and narrow chimney leads up from a fireplace; although you might be able to get up it, it seems unlikely you could get back down. Loosely attached to a wall is a small piece of paper.
Look: Studio This appears to have been an artist's studio. The walls and floors are splattered with paints of 69 different colors. Strangely enough, nothing of value is hanging here. At the south end of the room is an open door (also covered with paint). A dark and narrow chimney leads up from a fireplace; although you might be able to get up it, it seems unlikely you could get back down. Loosely attached to a wall is a small piece of paper.
Inv: You are carrying:   A painting   A brass lantern (providing light)
State Value: 6.91381
Top 20 Templates: up (0.99828) | open OBJ (0.00144) | east (0.00019) | close OBJ (1e-05) | open OBJ with OBJ (1e-05) | put OBJ in OBJ (1e-05) | north (1e-05) | turn on OBJ (1e-05) | south (1e-05) | shake OBJ (0.0) | put out OBJ (0.0) | pull OBJ (0.0) | west (0.0) | take OBJ (0.0) | take on OBJ (0.0) | put down OBJ (0.0) | go around OBJ (0.0) | southwest (0.0) | take all (0.0) | throw OBJ at OBJ (0.0)
Q Value for Valid Action: put out light (5.99698, 6.19877) | take paper (4.95, 5.08403) | up (7.9835, 8.05066) | south (6.35651, 6.22978) | put down light (5.93582, 6.13692) | put down painting (5.17499, 5.26189) | throw light at chimney (-3.43639, -3.62603)
Action: (up, knife, small) => up (Q Value: 7.9835)
Reward: 0
Score: 39
If Valid: True

=== STEP 17 ===
Game: Kitchen On the table is an elongated brown sack, smelling of hot peppers. A bottle is sitting on the table. The glass bottle contains:   A quantity of water
Look: Kitchen You are in the kitchen of the white house. A table seems to have been used recently for the preparation of food. A passage leads to the west and a dark staircase can be seen leading upward. A dark chimney leads down and to the east is a small window which is open. On the table is an elongated brown sack, smelling of hot peppers. A bottle is sitting on the table. The glass bottle contains:   A quantity of water
Inv: You are carrying:   A painting   A brass lantern (providing light)
State Value: 8.08167
Top 20 Templates: west (0.98782) | open OBJ (0.00466) | close OBJ (0.00219) | put out OBJ (0.00156) | east (0.00136) | turn on OBJ (0.00112) | south (0.00029) | down (0.00022) | throw OBJ at OBJ (0.00015) | north (0.00012) | put OBJ in OBJ (0.00011) | southwest (6e-05) | shake OBJ (6e-05) | blow out OBJ (4e-05) | wait (4e-05) | take OBJ (2e-05) | go around OBJ (2e-05) | put OBJ on OBJ (1e-05) | take on OBJ (1e-05) | pull OBJ (1e-05)
Q Value for Valid Action: jump (-9.27409, -9.04089) | east (6.91065, 7.0756) | open sack (7.64896, 7.67116) | open bottle (7.83443, 7.84561) | put out light (7.31911, 7.51925) | take sack (7.14957, 7.31653) | take bottle (7.36058, 7.56828) | up (7.63525, 6.93375) | close window (7.11225, 7.38313) | put down light (7.12197, 7.27129) | put down painting (6.84016, 6.65874) | put light on kitchen (7.48657, 7.47653) | put painting on kitchen (6.61161, 6.87845) | throw light at sack (7.20049, 7.30371) | west (8.42195, 8.46763) | take all (7.16313, 6.90915)
Action: (west, hide, hide) => west (Q Value: 8.42195)
Reward: 0
Score: 39
If Valid: True

=== STEP 18 ===
Game: Living Room Above the trophy case hangs an elvish sword of great antiquity.
Look: Living Room You are in the living room. There is a doorway to the east, a wooden door with strange gothic lettering to the west, which appears to be nailed shut, a trophy case, and a closed trap door at your feet. Above the trophy case hangs an elvish sword of great antiquity.
Inv: You are carrying:   A painting   A brass lantern (providing light)
State Value: 9.0133
Top 20 Templates: open OBJ (0.7769) | put OBJ in OBJ (0.14552) | throw OBJ at OBJ (0.02372) | take OBJ (0.0147) | turn on OBJ (0.01006) | put out OBJ (0.00705) | close OBJ (0.00569) | take all (0.00532) | fill OBJ with OBJ (0.00517) | take OBJ from OBJ (0.0036) | blow out OBJ (0.00098) | shake OBJ (0.00059) | open OBJ with OBJ (0.00031) | wait (0.00021) | put down OBJ (4e-05) | southwest (4e-05) | take on OBJ (3e-05) | east (2e-05) | north (1e-05) | west (1e-05)
Q Value for Valid Action: open case (9.32502, 9.38244) | open trap (8.79187, 8.83345) | put out light (8.73883, 8.67126) | take elvish (8.82231, 8.78069) | put down light (8.5772, 8.49365) | put down painting (8.12995, 8.15679) | throw light at case (8.76539, 8.70428) | east (7.8757, 7.77191)
Action: (open OBJ, case, sand) => open case (Q Value: 9.32502)
Reward: 0
Score: 39
If Valid: True

=== STEP 19 ===
Game: Opened.
Look: Living Room You are in the living room. There is a doorway to the east, a wooden door with strange gothic lettering to the west, which appears to be nailed shut, a trophy case, and a closed trap door at your feet. Above the trophy case hangs an elvish sword of great antiquity.
Inv: You are carrying:   A painting   A brass lantern (providing light)
State Value: 9.75245
Top 20 Templates: put OBJ in OBJ (0.99558) | take OBJ (0.00076) | throw OBJ at OBJ (0.00076) | turn on OBJ (0.00065) | fill OBJ with OBJ (0.00055) | open OBJ (0.00044) | put out OBJ (0.00031) | take OBJ from OBJ (0.0003) | close OBJ (0.00028) | open OBJ with OBJ (8e-05) | shake OBJ (7e-05) | take all (6e-05) | blow out OBJ (4e-05) | wait (2e-05) | southwest (2e-05) | south (2e-05) | pull OBJ (1e-05) | take on OBJ (1e-05) | east (0.0) | down (0.0)
Q Value for Valid Action: open trap (8.81373, 8.86289) | put out light (8.8782, 8.67625) | take elvish (8.62527, 8.54246) | close case (8.74766, 8.52331) | put down light (8.65828, 8.34951) | put down painting (7.95177, 7.87316) | put light in case (9.44403, 9.263) | put painting in case (10.24359, 10.13002) | throw light at case (8.95971, 8.86678) | east (8.19995, 8.14608)
Action: (put OBJ in OBJ, painti, case) => put painti in case (Q Value: 10.13002)
Reward: 6
Score: 45
If Valid: True

=== STEP 20 ===
Game: Done.
Look: Living Room You are in the living room. There is a doorway to the east, a wooden door with strange gothic lettering to the west, which appears to be nailed shut, a trophy case, and a closed trap door at your feet. Above the trophy case hangs an elvish sword of great antiquity. Your collection of treasures consists of:   A painting
Inv: You are carrying:   A brass lantern (providing light)
State Value: 4.1275
Top 20 Templates: east (0.87788) | put down OBJ (0.03258) | throw OBJ at OBJ (0.02263) | put OBJ in OBJ (0.02124) | close OBJ (0.02033) | fill OBJ with OBJ (0.00495) | open OBJ (0.0046) | put out OBJ (0.00398) | south (0.00307) | put OBJ on OBJ (0.00201) | blow out OBJ (0.00194) | take OBJ from OBJ (0.00118) | shake OBJ (0.00094) | north (0.00085) | turn on OBJ (0.00083) | wait (0.00027) | take all (0.00014) | take on OBJ (0.0001) | west (9e-05) | go around OBJ (9e-05)
Q Value for Valid Action: open trap (3.64628, 3.76126) | put out light (3.96223, 4.07976) | take painting (2.69356, 2.93009) | take elvish (3.79952, 3.76239) | close case (3.87295, 3.87445) | put down light (3.8273, 3.77032) | put light in case (4.00656, 4.07758) | throw light at case (3.95661, 3.94933) | east (4.58662, 4.68062)
Action: (east, air, hide) => east (Q Value: 4.58662)
Reward: 0
Score: 45
If Valid: True

=== STEP 21 ===
Game: Kitchen On the table is an elongated brown sack, smelling of hot peppers. A bottle is sitting on the table. The glass bottle contains:   A quantity of water
Look: Kitchen You are in the kitchen of the white house. A table seems to have been used recently for the preparation of food. A passage leads to the west and a dark staircase can be seen leading upward. A dark chimney leads down and to the east is a small window which is open. On the table is an elongated brown sack, smelling of hot peppers. A bottle is sitting on the table. The glass bottle contains:   A quantity of water
Inv: You are carrying:   A brass lantern (providing light)
State Value: 4.85421
Top 20 Templates: east (0.89858) | west (0.02883) | open OBJ (0.01785) | put OBJ on OBJ (0.01147) | throw OBJ at OBJ (0.01106) | put down OBJ (0.00712) | close OBJ (0.00674) | put out OBJ (0.00418) | south (0.00394) | take OBJ (0.00216) | take all (0.00215) | turn on OBJ (0.00134) | blow out OBJ (0.00111) | shake OBJ (0.00094) | put OBJ in OBJ (0.00083) | go around OBJ (0.00028) | wait (0.00026) | fill OBJ with OBJ (0.00023) | north (0.00012) | take OBJ from OBJ (0.00011)
Q Value for Valid Action: jump (-9.62053, -9.68663) | east (5.47266, 5.42416) | open sack (4.68312, 4.63459) | open bottle (4.71624, 4.69588) | put out light (4.56687, 4.65549) | take sack (4.53771, 4.46738) | take bottle (4.54206, 4.48889) | up (3.6283, 3.40506) | close window (4.4571, 4.26242) | put down light (4.47938, 4.51235) | put light on kitchen (4.58869, 4.6532) | throw light at sack (4.73553, 4.8077) | west (3.75647, 3.66998) | take all (4.45235, 4.43336)
Action: (east, air, hide) => east (Q Value: 5.42416)
Reward: 0
Score: 45
If Valid: True

=== STEP 22 ===
Game: Behind House
Look: Behind House You are behind the white house. A path leads into the forest to the east. In one corner of the house there is a small window which is open.
Inv: You are carrying:   A brass lantern (providing light)
State Value: 5.51456
Top 20 Templates: north (0.94676) | close OBJ (0.0175) | open OBJ (0.0127) | east (0.00788) | put down OBJ (0.00658) | put out OBJ (0.00189) | put OBJ in OBJ (0.0013) | turn on OBJ (0.0013) | throw OBJ at OBJ (0.00111) | south (0.00097) | shake OBJ (0.00045) | blow out OBJ (0.00035) | wait (0.00032) | fill OBJ with OBJ (0.00024) | southwest (0.00017) | put OBJ on OBJ (0.00011) | west (7e-05) | look around OBJ (6e-05) | take OBJ from OBJ (5e-05) | take all (4e-05)
Q Value for Valid Action: west (4.36668, 4.59523) | put out lantern (4.92282, 5.16851) | close window (5.09678, 5.0582) | put down lantern (4.97033, 5.2489) | south (4.79295, 5.00373) | throw lantern at white (5.33473, 5.40109) | north (5.69647, 5.76227) | east (4.95817, 5.09619)
Action: (north, air, air) => north (Q Value: 5.69647)
Reward: 0
Score: 45
If Valid: True

=== STEP 23 ===
Game: North of House You are facing the north side of a white house. There is no door here, and all the windows are boarded up. To the north a narrow path winds through the trees.
Look: North of House You are facing the north side of a white house. There is no door here, and all the windows are boarded up. To the north a narrow path winds through the trees.
Inv: You are carrying:   A brass lantern (providing light)
State Value: 6.26809
Top 20 Templates: north (0.98634) | open OBJ (0.00762) | up (0.00174) | south (0.00087) | close OBJ (0.00064) | put down OBJ (0.00048) | west (0.0004) | turn on OBJ (0.00035) | east (0.0003) | southwest (0.00027) | put out OBJ (0.00024) | take OBJ (0.00022) | go around OBJ (0.00015) | shake OBJ (0.0001) | wait (8e-05) | pull OBJ (6e-05) | put OBJ in OBJ (2e-05) | take on OBJ (2e-05) | down (2e-05) | northwest (2e-05)
Q Value for Valid Action: put out lantern (5.59677, 5.60613) | put down lantern (5.92268, 5.85248) | east (5.67092, 5.62873) | throw lantern at windows (5.76761, 5.76261) | north (6.52823, 6.49014) | west (5.11816, 5.35731)
Action: (north, egg, art) => north (Q Value: 6.49014)
Reward: 0
Score: 45
If Valid: True

=== STEP 24 ===
Game: Forest Path This is a path winding through a dimly lit forest. The path heads north south here. One particularly large tree with some low branches stands at the edge of the path.
Look: Forest Path This is a path winding through a dimly lit forest. The path heads north south here. One particularly large tree with some low branches stands at the edge of the path.
Inv: You are carrying:   A brass lantern (providing light)
State Value: 7.11361
Top 20 Templates: up (0.99878) | take OBJ (0.00032) | open OBJ (0.00031) | east (0.00026) | take all (0.00014) | put down OBJ (7e-05) | close OBJ (3e-05) | south (2e-05) | north (1e-05) | shake OBJ (1e-05) | go around OBJ (1e-05) | west (1e-05) | put out OBJ (1e-05) | turn on OBJ (0.0) | open OBJ with OBJ (0.0) | put OBJ in OBJ (0.0) | wait (0.0) | pull OBJ (0.0) | take on OBJ (0.0) | northwest (0.0)
Q Value for Valid Action: put out lantern (5.79732, 5.7651) | up (7.44611, 7.44857) | put down lantern (5.94259, 5.87923) | go around forest (5.52419, 5.53839) | throw lantern at large (5.72662, 5.95856) | north (5.20977, 5.45428) | south (5.57516, 5.72747) | west (5.37536, 5.81131) | east (5.46032, 5.38635)
Action: (up, egg, sand) => up (Q Value: 7.44611)
Reward: 0
Score: 45
If Valid: True

=== STEP 25 ===
Game: Up a Tree You are about 10 feet above the ground nestled among some large branches. The nearest branch above you is above your reach. Beside you on the branch is a small bird's nest. In the bird's nest is a large egg encrusted with precious jewels, apparently scavenged by a childless songbird. The egg is covered with fine gold inlay, and ornamented in lapis lazuli and mother of pearl. Unlike most eggs, this one is hinged and closed with a delicate looking clasp. The egg appears extremely fragile.
Look: Up a Tree You are about 10 feet above the ground nestled among some large branches. The nearest branch above you is above your reach. Beside you on the branch is a small bird's nest. In the bird's nest is a large egg encrusted with precious jewels, apparently scavenged by a childless songbird. The egg is covered with fine gold inlay, and ornamented in lapis lazuli and mother of pearl. Unlike most eggs, this one is hinged and closed with a delicate looking clasp. The egg appears extremely fragile.
Inv: You are carrying:   A brass lantern (providing light) You hear in the distance the chirping of a song bird.
State Value: 7.64377
Top 20 Templates: take OBJ (0.97704) | open OBJ (0.01301) | open OBJ with OBJ (0.00843) | take all (0.00129) | put down OBJ (6e-05) | put OBJ in OBJ (4e-05) | up (3e-05) | shake OBJ (3e-05) | turn on OBJ (2e-05) | put out OBJ (1e-05) | down (1e-05) | close OBJ (1e-05) | take on OBJ (1e-05) | west (0.0) | pull OBJ (0.0) | north (0.0) | south (0.0) | throw OBJ at OBJ (0.0) | go around OBJ (0.0) | northwest (0.0)
Q Value for Valid Action: down (6.57215, 6.45362) | put out brass (7.39907, 7.4741) | take egg (8.6295, 8.57796) | take nest (5.21378, 5.18105) | take on egg (5.4909, 5.42137) | close nest (5.88612, 5.75772) | put down brass (7.42487, 7.28912) | throw brass at egg (5.35724, 4.80761) | throw brass at branch (7.52678, 7.28126) | open egg with brass (7.92433, 7.82024)
Action: (take OBJ, egg, glue) => take egg (Q Value: 8.57796)
Reward: 5
Score: 50
If Valid: True

=== STEP 26 ===
Game: Taken.
Look: Up a Tree You are about 10 feet above the ground nestled among some large branches. The nearest branch above you is above your reach. Beside you on the branch is a small bird's nest.
Inv: You are carrying:   A jewel encrusted egg   A brass lantern (providing light)
State Value: 3.23792
Top 20 Templates: throw OBJ at OBJ (0.24127) | put down OBJ (0.23705) | open OBJ (0.08609) | down (0.07982) | take OBJ (0.06659) | put OBJ in OBJ (0.05395) | close OBJ (0.04744) | east (0.03921) | take on OBJ (0.02734) | shake OBJ (0.02168) | south (0.01782) | go around OBJ (0.01752) | open OBJ with OBJ (0.01595) | take all (0.00942) | pull OBJ (0.00803) | put out OBJ (0.00789) | turn on OBJ (0.00654) | up (0.00524) | west (0.00398) | blow out OBJ (0.00196)
Q Value for Valid Action: down (3.24215, 3.02172) | put out light (2.98628, 3.09725) | take nest (3.0036, 3.08614) | take on egg (3.05026, 3.14332) | close nest (3.01146, 3.10443) | put down egg (3.3905, 3.45777) | put down light (3.2267, 3.36651) | put egg in nest (3.3083, 3.33924) | put light in nest (3.25387, 3.33474) | throw egg at branch (3.29618, 3.44899) | throw egg at light (3.20545, 3.42111) | throw light at egg (3.34589, 3.35402) | open egg with light (3.01235, 3.2196)
Action: (east, glue, sand) => east (Q Value: 3.1457)
Reward: 0
Score: 50
If Valid: False

=== STEP 27 ===
Game: You can't go that way. You hear in the distance the chirping of a song bird.
Look: Up a Tree You are about 10 feet above the ground nestled among some large branches. The nearest branch above you is above your reach. Beside you on the branch is a small bird's nest.
Inv: You are carrying:   A jewel encrusted egg   A brass lantern (providing light)
State Value: 3.20465
Top 20 Templates: put down OBJ (0.24705) | throw OBJ at OBJ (0.2344) | open OBJ (0.07565) | take OBJ (0.07186) | east (0.05611) | down (0.05449) | close OBJ (0.05023) | put OBJ in OBJ (0.04801) | take on OBJ (0.02525) | go around OBJ (0.02374) | south (0.02338) | shake OBJ (0.0214) | open OBJ with OBJ (0.01293) | take all (0.01025) | pull OBJ (0.0088) | put out OBJ (0.00806) | west (0.00684) | up (0.00661) | turn on OBJ (0.00633) | north (0.00238)
Q Value for Valid Action: down (3.15664, 2.94927) | put out light (2.9687, 3.04238) | take nest (2.99218, 3.06096) | take on egg (3.05466, 3.10881) | close nest (3.00178, 3.06824) | put down egg (3.3461, 3.38145) | put down light (3.18467, 3.32081) | put egg in nest (3.2612, 3.26496) | put light in nest (3.20985, 3.28183) | throw egg at branch (3.20484, 3.35507) | throw egg at light (3.13902, 3.36796) | throw light at egg (3.30418, 3.33708) | open egg with light (2.99023, 3.14352)
Action: (take OBJ, painti, lanter) => take painti (Q Value: 3.19583)
Reward: 0
Score: 50
If Valid: False

=== STEP 28 ===
Game: You can't see any painti here!
Look: Up a Tree You are about 10 feet above the ground nestled among some large branches. The nearest branch above you is above your reach. Beside you on the branch is a small bird's nest.
Inv: You are carrying:   A jewel encrusted egg   A brass lantern (providing light)
State Value: 3.23488
Top 20 Templates: put down OBJ (0.24296) | throw OBJ at OBJ (0.23832) | open OBJ (0.07899) | down (0.07484) | take OBJ (0.06809) | put OBJ in OBJ (0.05249) | close OBJ (0.04828) | east (0.04427) | take on OBJ (0.02675) | shake OBJ (0.02151) | south (0.01954) | go around OBJ (0.0193) | open OBJ with OBJ (0.01486) | take all (0.00952) | pull OBJ (0.00837) | put out OBJ (0.00785) | turn on OBJ (0.00641) | up (0.00554) | west (0.00471) | blow out OBJ (0.00199)
Q Value for Valid Action: down (3.24046, 3.0271) | put out light (2.98764, 3.08322) | take nest (3.01897, 3.09299) | take on egg (3.07423, 3.1387) | close nest (3.02543, 3.10458) | put down egg (3.36704, 3.41842) | put down light (3.21498, 3.36032) | put egg in nest (3.2868, 3.30268) | put light in nest (3.24203, 3.31902) | throw egg at branch (3.24945, 3.40508) | throw egg at light (3.18896, 3.41085) | throw light at egg (3.34766, 3.35698) | open egg with light (3.01896, 3.18213)
Action: (put down OBJ, egg, egg) => put down egg (Q Value: 3.36704)
Reward: 0
Score: 50
If Valid: True

=== STEP 29 ===
Game: The egg falls to the ground and springs open, seriously damaged. There is a golden clockwork canary nestled in the egg. It seems to have recently had a bad experience. The mountings for its jewel like eyes are empty, and its silver beak is crumpled. Through a cracked crystal window below its left wing you can see the remains of intricate machinery. It is not clear what result winding it would have, as the mainspring seems sprung.
Look: Up a Tree You are about 10 feet above the ground nestled among some large branches. The nearest branch above you is above your reach. Beside you on the branch is a small bird's nest.
Inv: You are carrying:   A brass lantern (providing light) You hear in the distance the chirping of a song bird.
State Value: 3.58964
Top 20 Templates: throw OBJ at OBJ (0.23399) | put down OBJ (0.21573) | down (0.1423) | open OBJ (0.07503) | put OBJ in OBJ (0.05963) | take OBJ (0.05509) | close OBJ (0.04707) | east (0.03111) | take on OBJ (0.02938) | shake OBJ (0.02106) | open OBJ with OBJ (0.01745) | south (0.01478) | go around OBJ (0.01411) | pull OBJ (0.00795) | take all (0.00768) | put out OBJ (0.00747) | turn on OBJ (0.00674) | up (0.00432) | west (0.00297) | blow out OBJ (0.00169)
Q Value for Valid Action: down (3.88843, 3.83978) | put out brass (3.47859, 3.56044) | take nest (3.3801, 3.39584) | close nest (3.43626, 3.50913) | put down brass (3.64966, 3.71942) | put brass in nest (3.73879, 3.70496) | throw brass at branch (4.0722, 3.94825)
Action: (put OBJ in OBJ, sand, sand) => put sand in sand (Q Value: 3.41833)
Reward: 0
Score: 50
If Valid: False
\end{lstlisting}